\documentclass[lettersize,journal]{IEEEtran}
\usepackage{amsmath,amsfonts}
\usepackage{algorithmic}
\usepackage{algorithm}
\usepackage{array}
\usepackage[caption=false,font=normalsize,labelfont=sf,textfont=sf]{subfig}
\usepackage{textcomp}
\usepackage{stfloats}
\usepackage{url}
\usepackage{verbatim}
\usepackage{graphicx}
\usepackage{cite}
\usepackage{xcolor}

\usepackage{multirow}

\usepackage{bbding}

\begin{document}

\title{FedFM: Anchor-based Feature Matching for Data Heterogeneity in Federated Learning}
\author{Rui Ye, Zhenyang Ni, Chenxin Xu, Jianyu Wang, Siheng Chen~\IEEEmembership{Member,~IEEE}, Yonina C. Eldar~\IEEEmembership{Fellow,~IEEE}
\thanks{R. Ye, Z. Ni, C. Xu are with the Cooperative Medianet Innovation Center (CMIC) at Shanghai Jiao Tong University, Shanghai, China. 
E-mail: yr991129, 0107nzy, xcxwakaka@sjtu.edu.cn.}
\thanks {J. Wang is with Meta Platforms. E-mail: jianyuwang@meta.com.}
\thanks {S. Chen is with Shanghai Jiao Tong University and Shanghai AI laboratory, Shanghai, China, E-mail: sihengc@sjtu.edu.cn.}
\thanks {Y. C. Eldar is with Department of Computer Science and Applied Mathematics, Weizmann Institute of Science. E-mail: yonina.eldar@weizmann.ac.il.}

}


\maketitle

\begin{abstract}
One of the key challenges in federated learning (FL) is local data distribution heterogeneity across clients, which may cause inconsistent feature spaces across clients. To address this issue, we propose a novel method FedFM, which guides each client's features to match shared category-wise anchors (landmarks in feature space). This method attempts to mitigate the negative effects of data heterogeneity in FL by aligning each client's feature space. Besides, we tackle the challenge of varying objective function and provide convergence guarantee for FedFM. In FedFM, to mitigate the phenomenon of overlapping feature spaces across categories and enhance the effectiveness of feature matching, we further propose a more precise and effective feature matching loss called contrastive-guiding (CG), which guides each local feature to match with the corresponding anchor while keeping away from non-corresponding anchors. Additionally, to achieve higher efficiency and flexibility, we propose a FedFM variant, called FedFM-Lite, where clients communicate with server with fewer synchronization times and communication bandwidth costs. Through extensive experiments, we demonstrate that FedFM with CG outperforms several works by quantitative and qualitative comparisons. FedFM-Lite can achieve better performance than state-of-the-art methods with five to ten times less communication costs.
\end{abstract}


\section{Introduction}
\IEEEPARstart{M}{ost} existing deep learning models are trained in a centralized manner. However, in practice, data may be distributed on several parties and may not be collected due to the increasing privacy concerns. Federated learning (FL)~\cite{fedavg} is proposed to address this issue and has become an emerging research topic~\cite{gafni2022federated,advances,li_survey,fedavg,smith2017federated,fedprox,scaffold}. In standard FL~\cite{fedavg}, each client first downloads the same global model from the server and conducts local model training on its private dataset. Then, clients upload their trained local models to the server, where a global model is updated via local models aggregation. This process is conducted iteratively to obtain a final global model. This privacy-preserving method has been widely explored applied to many tasks, such as image classification~\cite{fedvisual,measuring}, language modeling~\cite{hard2018federated}, speech recognition~\cite{8683546}.

One of the key challenges that hinders FL from performing as well as centralized learning is data distribution heterogeneity across clients~\cite{abdulrahman2020survey,advances,wangsurvey}. Due to diverse conditions of devices and application scenarios, data might be not independent and identical distributed (IID) across local clients. This may result in large variations in the locally trained models on clients and slow down convergence of the global model~\cite{convergence_noniid,fednoniid}. This phenomenon is also referred to as \emph{client drift}~\cite{scaffold,wangsurvey}.

\begin{figure*}[!t]
\centering
\subfloat[FedAvg]{\includegraphics[width=2.1in]{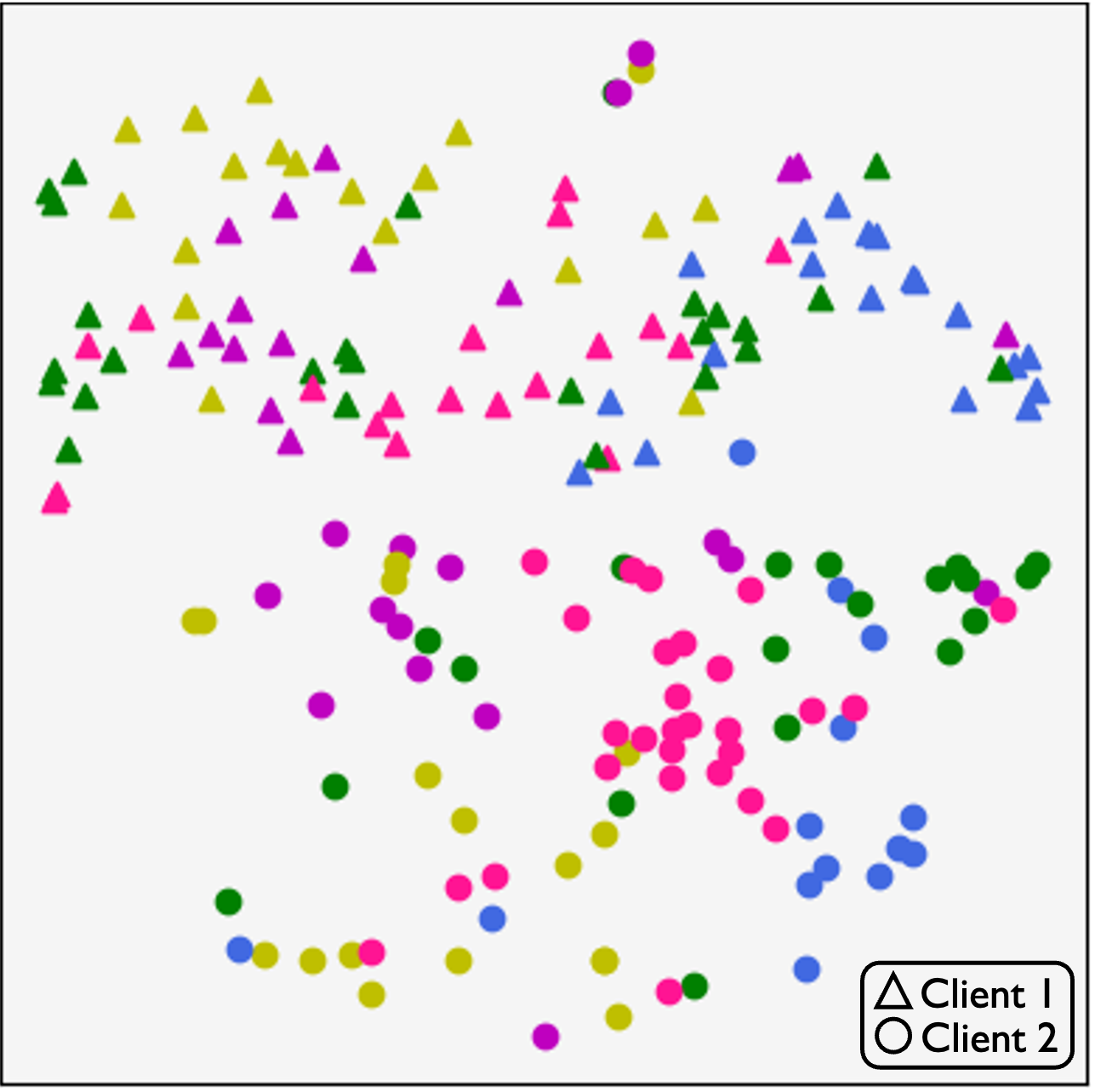}%
\label{fig:intro_avg}}
\hfil
\subfloat[FedFM with $\ell_2$ regularization]{\includegraphics[width=2.1in]{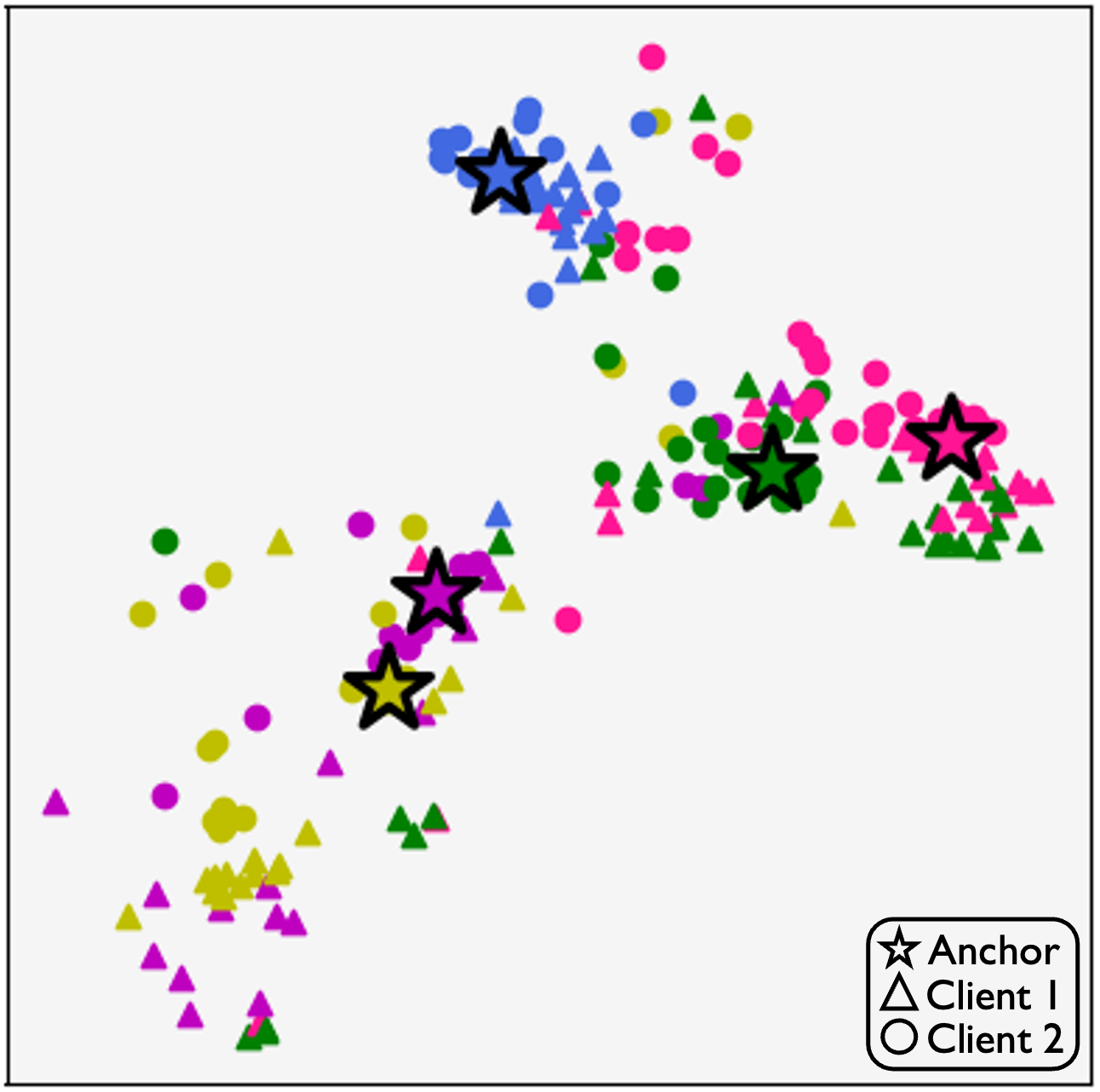}%
\label{fig:intro_fml2}}
\hfil
\subfloat[FedFM with CG]{\includegraphics[width=2.1in]{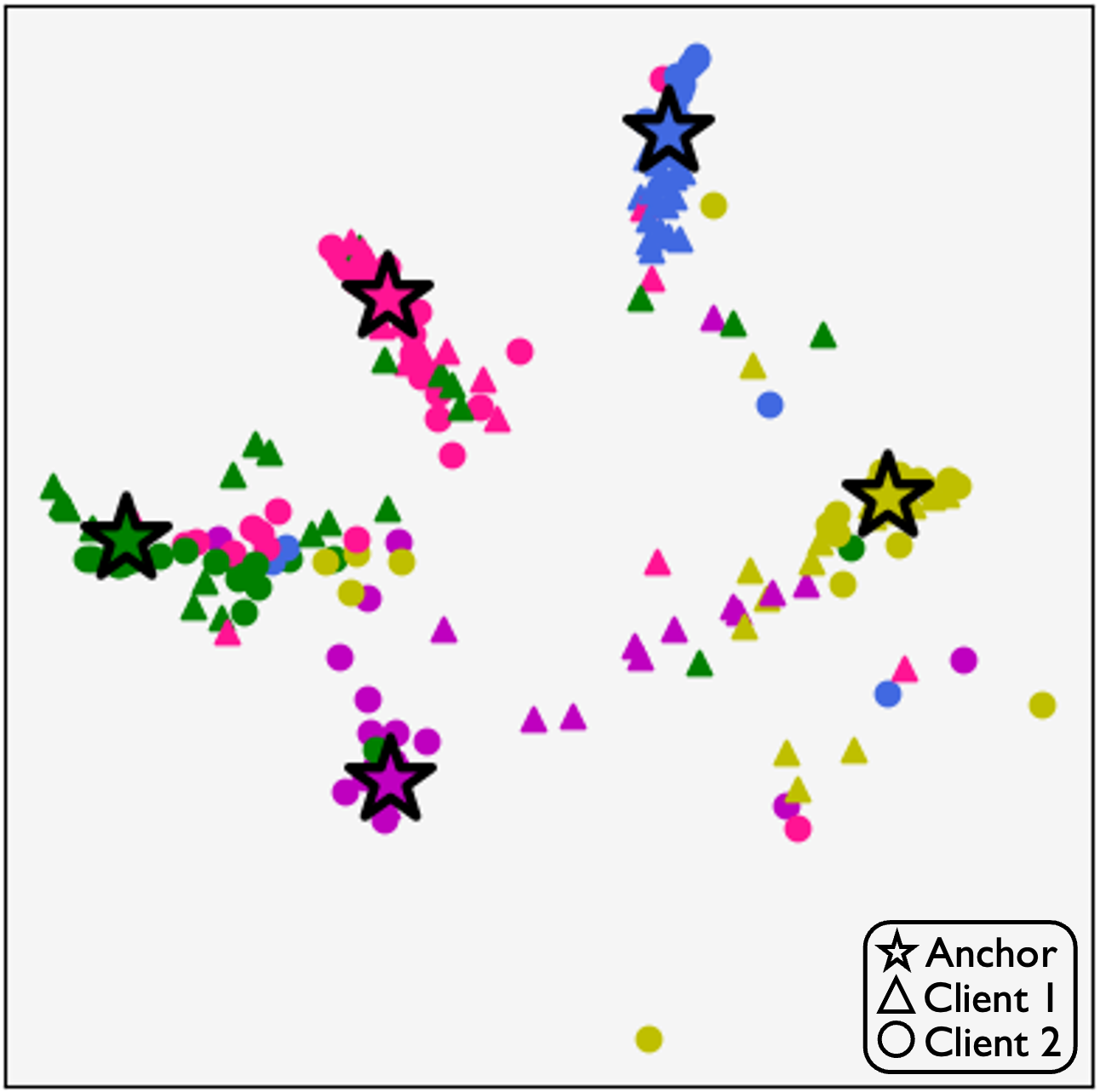}%
\label{fig:intro_fmcg}}
\caption{FedFM alleviates the inconsistency in feature space by anchor-based feature matching. (a) In existing methods, there is a large gap between samples of two clients (triangle and circle) in the feature space. (b) With simple $\ell_2$ regularization, our FedFM leverages anchors (stars) to align the feature spaces of two clients. (c) With the proposed contrastive-guiding (CG) method, FedFM achieves more precise and compact matching.}
\label{fig:intro}
\end{figure*}

To tackle the above mentioned data heterogeneity issue, most previous works~\cite{fedprox,scaffold,feddyn} focus on model-level corrections, which intend to reduce the variations in locally trained models. However, these methods fail to ensure the consistency of multiple local models' feature spaces. It is possible that different local models have drastically misaligned feature spaces. This could lead to unclear decision boundaries and cause misclassification, which significantly differs from centralized learning. Fig.~\ref{fig:intro_avg} empirically shows the T-SNE~\cite{tsne} of two local clients' features in FedAvg~\cite{fedavg}, where the color indicates categories and the shape indicates clients. We see that the data samples with the same color, yet different shapes do not overlap, reflecting two local models fail to share a consistent feature space. In addition, data samples with the same shape, yet different colors overlap with each other. This can be detrimental to classification tasks. Motivated by this, our work focuses on mitigating the data heterogeneity issue in federated classification tasks through aligning the feature spaces across multiple local models.

In this paper, we propose an anchor-based \textbf{Fed}erated \textbf{F}eature \textbf{M}atching (FedFM) method, the key idea of which is to leverage landmarks shared by all clients to provide global positioning, promoting a more consistent feature space. As a core concept of FedFM, we define landmarks as the average of features for the same class/category and name them as \emph{anchors}. In each round of FedFM, there are two key steps: (1) anchor updating; and (2) anchor-based model updating. In the anchor updating step, first, each client calculates the local anchors; second, by interacting with the server, global anchors are updated by aggregating local anchors and sent back to each client. In the anchor-based model updating step, each client's feature is pushed to match with the global anchor of the corresponding category during the process of local model training; see the significant improvement in Fig.~\ref{fig:intro_fml2}, where we regularize the $\ell_2$ distance between a feature and its corresponding global anchor to enhance the consistency of feature spaces across clients. Global anchors are denoted by star shape.

Moreover, we conduct theoretical analysis to provide convergence guarantee for our proposed FedFM algorithm. Unlike most existing literature that analyzes fixed objective functions, the analysis of FedFM faces a distinctive challenge of time-varying objective functions over rounds. This is because of the varying global anchors, which are updated at each round. We overcome this challenge by proving a key lemma, which suggests that updating of global anchors also contributes to optimize the global objective. The theoretical results show that the proposed FedFM converges at a rate that accords with many existing theoretical optimization literature~\cite{fednova,convergence_noniid}. 

To promote more precise and effective feature matching and push the feature spaces of different categories to be far away from each other, we further propose contrastive-guiding (CG) for feature matching in FedFM. The proposed CG guides each client's local feature to match with the corresponding global anchor while keeping away from non-corresponding global anchors. Comparing with the standard $\ell_2$ regularization, CG contributes to more precise feature matching, which results in more distant and compact category-wise feature space; see its improvement over $\ell_2$ regularization in Fig.~\ref{fig:intro_fmcg}.

To achieve higher efficiency and flexibility, we propose a variant of FedFM, called FedFM-Lite. Compared with FedFM, FedFM-Lite is more efficient since it communicates only one time within one federated round and thus requires less synchronization times (handshakes) among clients and server. FedFM-Lite is also more flexible to communicate anchors and models at different frequency. Since models have significantly more communication bandwidth cost, we propose to communicate models at a relatively lower frequency, which is capable to accord with various real-world communication budgets.

At last, through extensive experiments, we verify that FedFM with CG outperforms the state-of-the-art FL methods, including FedAvg~\cite{fedavg}, FedAvgM~\cite{fedavgm}, FedProx~\cite{fedprox}, SCAFFOLD~\cite{scaffold}, FedDyn~\cite{feddyn}, FedNova~\cite{fednova} and MOON~\cite{moon}, on multiple datasets, including CIFAR-10~\cite{cifar10}, CINIC-10~\cite{darlow2018cinic} and CIFAR-100. We further visualize the feature space constructed by the proposed FedFM with CG, which qualitatively demonstrates its effectiveness. We also see that FedFM-Lite can achieve better performance than existing methods with five to ten times less communication costs and comparable performance compared with FedFM with half of the synchonization times.

Our main contributions are as follows:
\begin{enumerate}
\item We propose an anchor-based federated feature matching (FedFM) method and a contrastive-guiding (CG) technique in FedFM, which pushes each client's local feature to match with corresponding shared global anchor while keeping away from non-corresponding anchors, promoting a consistent feature space across clients and mitigating the notorious data heterogeneity issue;

\item We tackle the distinctive challenge of varying objective function in the theoretical analysis of FedFM and provide a convergence guarantee;

\item We propose an efficient and flexible variant, FedFM-Lite, which can be easily adjusted to accord with various real-world communication budgets;

\item We conduct extensive experiments and show that FedFM with CG (and FedFM-Lite) can significantly outperform state-of-the-art methods.
\end{enumerate}

This paper is organized as the following. Section~\ref{sec:related_work} reviews related works. Section~\ref{sec:preliminaries} presents several preliminaries including notations and motivations. Section~\ref{sec:methodology} describes and discusses our proposed FedFM method and the CG technique. Section~\ref{sec:convergence_analysis} provides convergence analysis for FedFM. Section~\ref{sec:fedfm_lite} proposes a variant of FedFM, FedFM-Lite. Section~\ref{sec:experiments} shows the experimental results.

\section{Related Work}
\label{sec:related_work}

Federated learning (FL) is proposed in~\cite{fedavg} and has been widely applied to many fields. In image processing, it is widely adopted since it takes advantage of the computational ability and locally-stored data of edge devices~\cite{fedvisual,measuring}. It also attracts much attention in healthcare due to its privacy-preserving property~\cite{fedmedical,fedhealth,9832948}. But when the standard FL method FedAvg~\cite{fedavg} meets the situation of data distribution heterogeneity across clients, the global model could move far away from the true global optimum due to the large variations in each local optima, which is referred to as client drift~\cite{scaffold}. There have been numerous works trying to tackle this issue and two key approaches are local correction and global adjustment.

\subsection{Local Correction}
One main approach is conducting correction during the process of local model training, which aims to reduce the difference among trained local models. Most previous works conduct correction on the \emph{model-level}. FedProx~\cite{fedprox} applies a $l_2$-norm distance regularization between the current local model and the previous global model. FedDyn~\cite{feddyn} proposes a dynamic regularizer to align local and global solutions. Variance reduction methods, such as SCAFFOLD~\cite{scaffold} and VRLSGD~\cite{vrlsgd}, utilize the previous difference between local and global gradient to debias the gradient at each local training step. MOON~\cite{moon} maximizes the similarity between the feature (intermediate layer output) of current local model and that of the previous global model, which requires three times computation cost and has no convergence guarantee. Our proposed FedFM conducts local correction in the~\emph{feature-level}, which employs shared category-wise global anchors to guide local feature learning. That is, MOON~\cite{moon} aligns features of two models that belong to the same sample and FedFM aligns features of all samples that belong to the same category. We also provide a convergence guarantee of FedFM.

\subsection{Global Adjustment}

Another key direction is global adjustment during the process of model interaction, which aims to obtain a better global model utilizing the uploaded local models. FedAvgM~\cite{fedavgm} introduces momentum to global model updating, which stabilizes the global model optimization. FedNova~\cite{fednova} normalizes local updates according to the number of SGD steps, which eliminates objective inconsistency and achieves fast convergence. Using the Knowledge-Distillation technique, FedGen~\cite{fedgen} learns a generator to assist local model training. FedDF~\cite{feddf} and FedFTG~\cite{fedftg} refine the global model by learning from the uploaded local models. Our proposed FedFM is orthogonal to these global adjustment methods and can be easily incorporated with these techniques.

As the above local correction and global adjustment methods, the focus of this paper is generalized FL, which aims for collaboratively training a global model. Personalized FL aims at collaboratively training multiple personalized local models, including FedRep~\cite{fedrep}, FedAMP~\cite{fedamp}, pFedMe~\cite{pfedme} and Personalized FedAvg~\cite{per_ml_2}. Targeting personalized FL, FedProto~\cite{tan2021fedproto} utilizes prototype to provide extra feature information from other clients to enhance personalization of each client. In comparison, our FedFM targets generalized FL, which uses anchors as landmarks to align clients' category-wise feature spaces to enhance generalization of all clients. We also propose a new contrastive-guiding (CG) technique. CG pushes local feature close to corresponding anchor and keeps it far away from non-corresponding anchors, which is shown to be significantly effective.

We compare FedFM with several representative methods in generalized FL in Table~\ref{table:related_work}.

\begin{table}[t]
\begin{center}
\caption{Related work comparisons. Conv. denotes convergence guarantee. Mem. and Band. denote memory cost and bandwidth cost roughly compared with FedAvg.}
\label{table:related_work}
\begin{tabular}{cccccc}
   \hline
   Method & Local Correction & Conv. & Mem. & Band.\\
   \hline
   FedAvg~\cite{fedavg}    & - & \checkmark & $\times$ 1 & $\times$ 1 \\
   FedAvgM~\cite{fedavgm}  & - & \checkmark & $\times$ 1 & $\times$ 1 \\
   FedProx~\cite{fedprox}  & Model & \checkmark & $\times$ 2 & $\times$ 1 \\
   SCAFFOLD~\cite{scaffold}& Model & \checkmark & $\times$ 2 & $\times$ 2 \\
   FedDyn~\cite{feddyn}    & Model & \checkmark & $\times$ 2 & $\times$ 1\\
   FedNova~\cite{fednova}  & - & \checkmark & $\times$ 1 & $\times$ 1\\
   MOON~\cite{moon}        & Feature & - & $\times$ 3 & $\times$ 1 \\
   FedFM (ours)            & Feature & \checkmark & $\times$ 1 & $\times$ 1 \\
   
   \hline
\end{tabular}
\end{center}
\end{table}

\section{Preliminaries}
\label{sec:preliminaries}
In this section, we present several key notations and the general process of FL. Then, we demonstrate two key empirical observations through preliminary experiments, including inconsistent feature spaces across clients and overlapping feature spaces across categories, which motivate the proposal of our FedFM method and CG loss.

\subsection{Notations}
Suppose there are $K$ clients, where the $k$th client holds a local dataset $\mathcal{B}_k=\{ (\mathbf{x}_i,c_i) | i=1,2,...,|\mathcal{B}_k|\}$, where $\mathbf{x}_i$ and $c_i$ are the data and the label of the $i$th sample, respectively. FL aims to leverage the local datasets at multiple clients to collectively train a global model $\mathbf{w}$ in the server without sharing raw data~\cite{fedavg}. Here we focus on a $C$-classification task and each local dataset $\mathcal{B}_k$ can be further split to $C$ category-wise sub-datasets, each of which is $\mathcal{B}_{k,c}=\{ (\mathbf{x}_i,c_i) \in \mathcal{B}_k | c_i=c \}$.  Let $f_{\mathrm{full}}(\cdot,\cdot)$ be an end-to-end classification model with $f_{\mathrm{full}}(\mathbf{w},\mathbf{x}) \in \mathbb{R}^C$ the final classification output given the input data sample $\mathbf{x}$ and model parameters $\mathbf{w}$. We also consider the intermediate features as $f_{\mathrm{mid}}(\mathbf{w},\mathbf{x}) \in \mathbb{R}^d$, where $f_{\mathrm{mid}}(\cdot,\cdot)$ denotes the feature-extract module in the full classification model $f_{\mathrm{full}}(\cdot,\cdot)$. A standard  global objective of FL is 
\begin{equation}
\label{eq:fedavg}
 F (\mathbf{w}) = \sum_{k=1}^K p_k F_k(\mathbf{w}) 
 = \sum_{k=1}^K \frac{p_k}{|\mathcal{B}_k|}
 \sum_{(\mathbf{x},c) \in \mathcal{B}_k} \ell( f_{\mathrm{full}}(\mathbf{w},\mathbf{x}),c),
\end{equation}
where $p_k={|\mathcal{B}_k|}/{\sum_{k=1}^{K}{|\mathcal{B}_k|}}$ is the aggregation weight of the $k$th client and $\ell(\cdot)$ is the task-specific loss function. To optimize the objective in a federated setting, at each communication round $t$, each client $k$ downloads the same global model $\mathbf{w}^{(t)}$ and conducts $\tau$ iterations of SGD on it, which normally applies supervision on final output $f_{\mathrm{full}}(\mathbf{w},\mathbf{x})$ while we could also consider supervision on the feature $f_{\mathrm{mid}}(\mathbf{w},\mathbf{x})$. Then, each client $k$ uploads the updated model $\mathbf{w}^{(t, \tau)}_k$ to the server, which is aggregated to update the global model $\mathbf{w}^{(t+1)}$ for next round.

 Since each local dataset $\mathcal{B}_k$ could have different data distributions, conventional method~\eqref{eq:fedavg} could results in divergent global model and degraded performance. In this work, our goal is to introduce a regularization term to the global objective, mitigating the effects of data heterogeneity.

\subsection{Motivation}

\label{ssct:me}

\begin{figure}[!t]
\centering
\subfloat[Inconsistent features]{\includegraphics[width=1.7in]{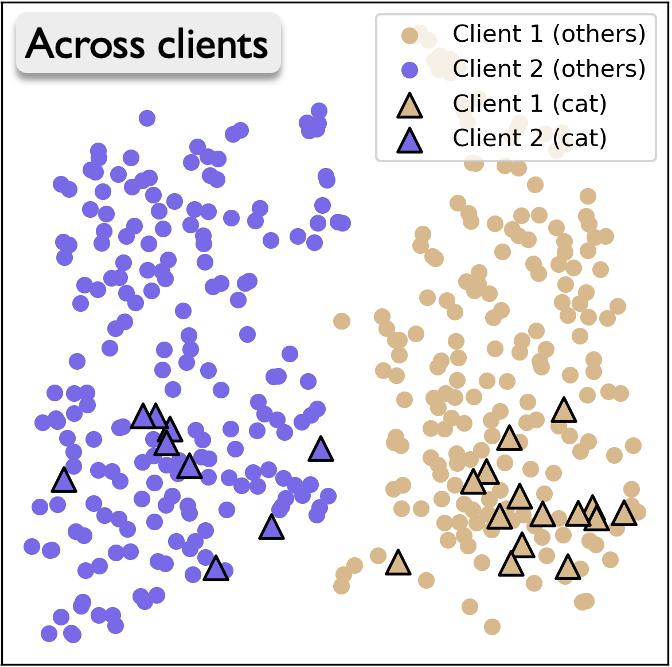}%
\label{fig:motivating_1}}
\hfil
\subfloat[Overlapping features]{\includegraphics[width=1.7in]{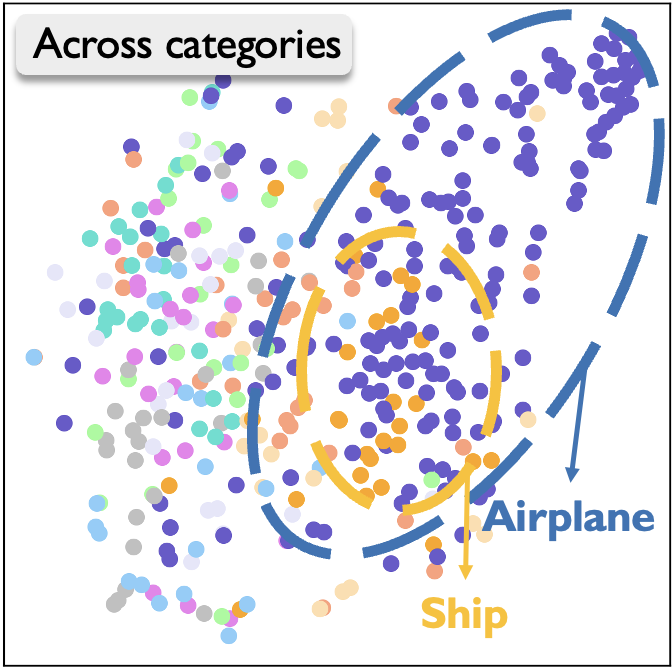}%
\label{fig:motivating_2}}
\caption{Motivating examples. Fig.~\ref{fig:motivating_1} visualizes the scatter plot of intermediate features of all the samples across two clients, where each color denotes one client. The triangles are the features of cat category. This shows that although each client has learned well-clustered features, the features are quite inconsistent across clients. Fig.~\ref{fig:motivating_2} shows the scatter plot of intermediate features of all samples in Client 1, while each color denotes one category. This shows that airplane category occupies larger feature space and overlap with that of ship category.
}
\label{fig:motivating}
\end{figure}

Most FL methods do not explore clients' behavior in feature space. We conduct the following FL (FedAvg~\cite{fedavg}) experiment on CIFAR-10 with two clients. Each holds an imbalanced dataset of size $5000$, where Client 1 has $50\%$ of data in airplane category while the rest is equally distributed to $9$ other categories (car category for Client 2). Each client runs $10$ epochs of local model training based on the same initial model. We use a ResNet18~\cite{resnet} model to implement $f_{\mathrm{full}}(\cdot,\cdot)$, where the feature-extraction module $f_{\mathrm{mid}}(\cdot,\cdot)$ is ResNet18 without the last fully-connected layer. We then visualize intermediate features of several validation sample by T-SNE~\cite{tsne}. From the experimental results, we notice two unsatisfying phenomena in feature space that might cause bad performance in FL.

\textbf{Inconsistent feature spaces across clients.} Fig.~\ref{fig:motivating_1} illustrates the scatter plot of several samples' features from two clients, where two colors indicate two different clients and cat category is highlighted in the triangle shape. We see that the samples from cat category across two local clients have a huge gap, reflecting that the feature spaces of two local clients are seriously inconsistent. This distinctly differs from centralized training, where samples of the same category should be gathered without an obvious gap. The intuition behind this phenomenon might be that two clients have two significantly different data distributions, resulting in distinct local models and therefore inconsistent behaviors in feature space. Motivated by this, we propose FedFM (anchor-based federated feature matching) to align the category-wise feature spaces across clients. The core idea is to establish shared global anchors as the landmarks in the feature space to guide feature learning across multiple clients; see details in Section~\ref{ssct:architecture}.

\textbf{Overlapping feature spaces across categories.} Fig.~\ref{fig:motivating_2} illustrates the scatter plot of several samples' features from various categories within Client 1, where various colors indicate different categories. We see that the samples from two categories (airplane and ship) greatly overlap, causing misclassification. This phenomenon often happens when the sample sizes across multiple categories are highly imbalanced. Motivated by this, we further propose a contrastive-guiding method in FedFM, which pushes each feature close to its corresponding anchor while keeping far away from non-corresponding anchors to avoid overlapping. This also ends up enlarging the distance between two distinct categories in the feature space and further mitigates overlapping; see details in Section~\ref{ssct:cg}.

\begin{figure*}[t]
\centering
\includegraphics[scale=1.0]{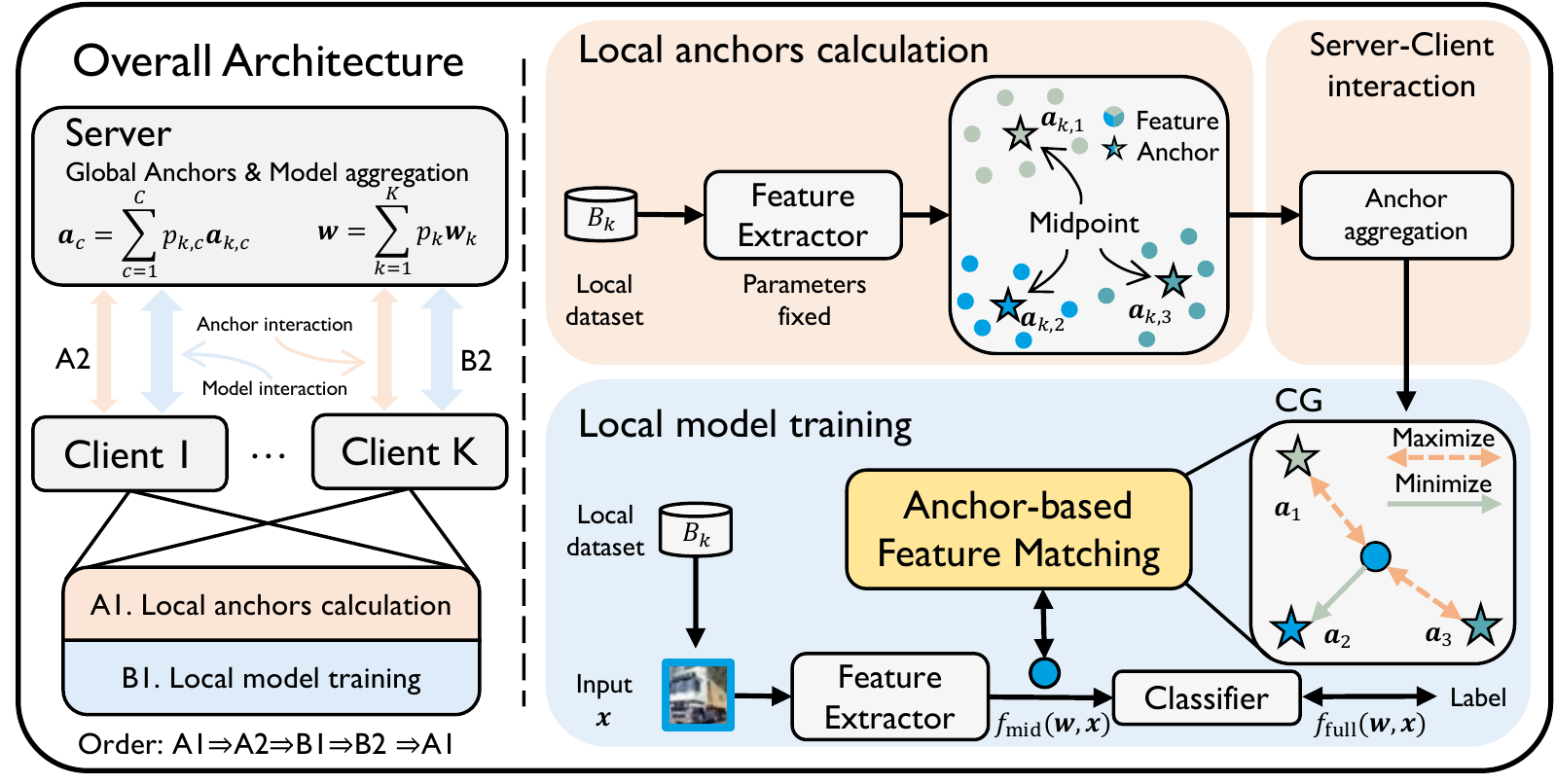}
\caption{Overview of FedFM for a 3-classification task. The left shows the overall architecture. The right shows two key steps in detail, where local anchors calculation generates feature anchors for each category and local model training utilizes these anchors to conduct anchor-based feature matching (e.g. contrastive guiding in the figure, which is described in Section~\ref{ssct:cg}). Here, a feature (in circle shape) is the intermediate layer output of a model and an anchor (in star shape) is an integration of features that belong to the same category.}
\label{fig:overview}
\end{figure*}

\section{Methodology}
\label{sec:methodology}
This section introduces the proposed federated learning with anchor-based feature matching (FedFM) from both aspects of mathematical optimization and federated implementation. Based on the proposed framework, we further propose a contrastive-guiding (CG) loss to mitigate overlapping feature spaces across categories. Finally, we discuss the communication cost and privacy concerns.

\subsection{Optimization Problem}
\label{sec:optimization_objective}
To address the issue of inconsistent feature spaces across clients, we propose anchor-based feature matching, which introduces anchors to serve as the shared landmarks for aligning all the clients' feature spaces. Mathematically, let $\mathcal{A}=\{ \mathbf{a}_c \}_{c=1}^C$ be a global anchor set, where $\mathbf{a}_c$ is the anchor of the $c$th category. The overall optimization problem with respect to the model parameter $\mathbf{w}$ and the anchor set $\mathcal{A}$ is
\begin{eqnarray}
\label{eq:global_objective}
 \mathop{\min}_{\mathbf{w}, \mathcal{A}} \Phi (\mathbf{w};\mathcal{A})
& = & \mathop{\min}_{\mathbf{w}, \mathcal{A}} \sum_{k=1}^K p_k \Phi_k(\mathbf{w};\mathcal{A})
\\ \nonumber
& = & \mathop{\min}_{\mathbf{w}, \mathcal{A}} \sum_{k=1}^K p_k \bigg( F_k(\mathbf{w})+\lambda Q_k(\mathbf{w};\mathcal{A}) \bigg),
\end{eqnarray}
where $p_k$ is the predefined aggregation weight of the $k$th client, with relative dataset size a standard choice ${|\mathcal{B}_k|}/{\sum_{k=1}^{K}{|\mathcal{B}_k|}}$, $\Phi_k(\cdot)$ is the $k$th client's objective,  $\lambda$ is a hyperparameter to balance the task-specific loss and the regularization term, 
$
F_k(\mathbf{w}) = \sum_{(\mathbf{x},c) \in \mathcal{B}_k}  \ell \left( f_{\mathrm{full}}(\mathbf{w},\mathbf{x}),c \right)/|\mathcal{B}_k| 
$
is the task-specific loss at the $k$th client with $\mathcal{B}_k$ the $k$th client's local dataset and 
\begin{eqnarray}
Q_k(\mathbf{w};\mathcal{A}) 
& = & \sum_{(\mathbf{x},c) \in \mathcal{B}_k} 
\frac{1}{{|\mathcal{B}_k|}} q(f_{\mathrm{mid}}(\mathbf{w},\mathbf{x}),\mathcal{A}|c)
 \nonumber \\  \label{eq:l2_term}
& = & \sum_{(\mathbf{x},c) \in \mathcal{B}_k} 
\frac{1}{{|\mathcal{B}_k|}} \left\| f_{\mathrm{mid}}(\mathbf{w},\mathbf{x}) - \mathbf{a}_c \right\|^2_2
\end{eqnarray}
is the $k$th client's anchor-based feature matching term, which forces each sample to match with the corresponding category-wise global anchor at each client. Since each global anchor is the proxy of each category and is shared across all the clients, with the anchor-based matching, the feature space at each client is evolving towards the same formation, enhancing the feature consistency across clients. Without anchor-based regularization term $Q_k(\mathbf{w};\mathcal{A})$, the overall objective degenerates to the standard federated learning objective.

To solve the optimization~\eqref{eq:global_objective}, we sequentially optimize the anchor set $\mathcal{A}$ and the model parameter $\mathbf{w}$ at each round $t$. 

\paragraph{Optimizing global anchors $\mathcal{A}$} Fixing the model parameter at the previous round, $\mathbf{w}^{(t)}$, we optimize over the the anchor set $\mathcal{A}$ by solving
\begin{equation}
\label{eq:global_anchors}
\mathcal{A}^{(t)} \ = \ \mathop{\arg\min}_{\mathcal{A}} \Phi (\mathbf{w}^{(t)};\mathcal{A})=\sum_{k=1}^K p_k \Phi_k(\mathbf{w}^{(t)};\mathcal{A}).
\end{equation}
Since the task-specific loss has nothing to do with the anchors, the optimal anchor only relates to the anchor-based regularization term. Furthermore, the global anchor of the $c$th category only depends on the data sample belonging to the $c$th category. Mathematically, the global anchor of the $c$th category has a straightforward closed-form solution as
\begin{eqnarray}
\mathbf{a}_c^{(t)} 
& = & \mathop{\arg\min}_{\mathbf{a}} \sum_{k=1}^K \sum_{(\mathbf{x},c) \in \mathcal{B}_{k}} \left\| f_{\mathrm{mid}}(\mathbf{w}^{(t)},\mathbf{x}) - \mathbf{a}_c \right\|^2_2 
\nonumber \\  \label{eq:global_anchors_compute}
& = &  \frac{1}{\sum_{k=1}^K |\mathcal{B}_{k,c}|} \sum_{k=1}^K  \sum_{(\mathbf{x},c) \in \mathcal{B}_{k,c}} f_{\mathrm{mid}}(\mathbf{w}^{(t)},\mathbf{x}),
\end{eqnarray}
where $\mathcal{B}_{k,c}$ is the $k$th client's local dataset that belongs to the $c$th category.  However, in the federated learning setting, since the global server cannot directly access the local data, each global anchor cannot be directly computed in the global server as in~\eqref{eq:global_anchors_compute}. In Section~\ref{ssct:architecture}, we will consider a federated implementation.

\paragraph{Optimizing global model $\mathbf{w}$} Fixing the global anchors $\mathcal{A}^{(t)}$,  we optimize over the the model parameter $\mathbf{w}$ by solving
\begin{equation}
\label{eq:global_model}
\mathbf{w}^{(t+1)} = \mathop{\arg\min}_{\mathbf{w}} \Phi (\mathbf{w};\mathcal{A}^{(t)})=\sum_{k=1}^K p_k \Phi_k(\mathbf{w};\mathcal{A}^{(t)}).
\end{equation}
We can consider an iterative solver based on the standard gradient descent. Mathematically, the global model parameters can be updated as
\begin{equation}
\label{eq:global_model_update}
\mathbf{w}^{(t+1)} = \mathbf{w}^{(t)} - \eta \sum_{k=1}^K p_k \bigg(  \frac{\partial F_k(\mathbf{w})}{\partial \mathbf{w}} + \lambda \frac{\partial  Q_k(\mathbf{w};\mathcal{A}^{(t)}) }{\partial \mathbf{w}} \bigg),
\end{equation}
where $\eta$ is the step size. Similarly to the optimization of global anchors, the model parameters cannot be directly updated in the global server as in~\eqref{eq:global_model_update}. A federated implementation is introduced in next.

\subsection{Federated Implementation}\label{ssct:architecture}
Previously, we propose the mathematical optimization of federated learning with anchor-based feature matching. However, the solver is impractical due to the federated setting. Here we introduce a federated implementation, called FedFM.

\subsubsection{Overview.}  Fig.~\ref{fig:overview} overviews the proposed FedFM. In each communication round, it consists of two main steps: anchor updating and model updating, where anchor updating solves the subproblem~\eqref{eq:global_anchors} and model updating solves the subproblem~\eqref{eq:global_model}.

In the step of \textbf{anchor updating}, after downloading the same global model, each client calculates local anchors and uploads them to the server, where global anchors are updated by aggregating local anchors. These global anchors are then broadcast to be shared by all clients. In the step of \textbf{model updating}, each client conducts several iterations of local model training supervised by task-driven loss as well as an anchor-based feature matching loss, after which the updated local model is uploaded to server. Then, the server updates the global model by aggregating local models.

\begin{algorithm}[t]
\caption{FedFM}
\label{alg:fedfm}
\begin{algorithmic}
\STATE \textbf{Initialization:} Global model $\mathbf{w}^{(0)}$.
\FOR{$t=0,1,...,T-1$}
\STATE Sends global model $\mathbf{w}^{(t)}$ to initialize each client $ \mathbf{w}_k^{(t,0)}$
\STATE $\mathcal{A}^{(t)}_k$ $\leftarrow$ \textbf{Local Anchors Calculation} ($\mathbf{w}_k^{(t,0)}$) using \eqref{eq:localp}
\STATE $\mathcal{A}^{(t)}$ $\leftarrow$ \textbf{Global Anchors Aggregation} ($\{ \mathcal{A}^{(t)}_k \}_{k=1}^K$) using \eqref{eq:globalp}
\STATE $\mathbf{w}^{(t,\tau)}_k$ $\leftarrow$ \textbf{Local Model Training} ($\mathbf{w}_k^{(t,0)}$,$\mathcal{A}^{(t)}$) for $\tau$ iterations using \eqref{eq:local_model_update}
\STATE $\mathbf{w}^{(t+1)}$ $\leftarrow$ \textbf{Global Model Aggregation} ($\{ \mathbf{w}^{(t,\tau)}_k \}_{k=1}^K$) using \eqref{eq:modelaggr}
\ENDFOR
\STATE \textbf{return} final global model $\mathbf{w}^{(T)}$
\end{algorithmic}
\end{algorithm}

We now illustrate these two steps in detail.

\paragraph{Step 1: Anchor Updating} Here we aim to implement~\eqref{eq:global_anchors_compute} in a federated fashion, which requires the coordination of both the clients and the server. This involves two substeps: local anchors calculation for integrating features of the same category, working on the client side, and global anchors aggregation for aggregating anchors from all clients, working on the server side.

\emph{Local anchors calculation.} After downloading the global model, at the start of each new round, each client conducts local anchors calculation by computing the category-wise midpoints of features; that is, local anchors are integration of features from the same category in each client. Mathematically, let $\mathbf{w}_k^{(t,r)}$ be the $k$th client's model parameter at the $t$th communication round with iteration $r$ and $\mathbf{w}_k^{(t,0)} := \mathbf{w}^{(t)}$, which means that the local client's model parameter at iteration $0$ is initialized by the global model in the previous communication round. Then, the local anchor of the category $c$ in client $k$ at round $t$ is calculated as
\begin{equation}
\label{eq:localp}
    \mathbf{a}^{(t)}_{k,c}=\frac{1}{|\mathcal{B}_{k,c}|} \sum_{(\mathbf{x},c) \in \mathcal{B}_{k,c}} f_{\mathrm{mid}}(\mathbf{w}^{(t)},\mathbf{x}) \in \mathbb{R}^d,
\end{equation}
where $f_{\mathrm{mid}}(\mathbf{w},\mathbf{x})$ is the intermediate layer output (feature) of a model $\mathbf{w}$ given a sample $\mathbf{x}$. Here $\mathbf{a}^{(t)}_{k,c} \in \mathbb{R}^d$ is essentially the midpoint of features of category $c$ data in sub-dataset $\mathcal{B}_{k,c}$. Client $k$ performs the calculation for each category $c$ and then sends all $C$ local anchors to the server.

\emph{Global anchors aggregation.} The server conducts global anchors aggregation by aggregating local anchors from clients so that global anchors are integration of features from the same category across all clients. Mathematically, receiving the local anchors from all the $K$ clients, the server conducts the following dataset size weighted aggregation for each category $c$ to obtain global anchor; that is,
\begin{equation}
\label{eq:globalp}
\begin{aligned}
    \mathbf{a}_c^{(t)}
    :&= \frac{1}{\sum_{k=1}^K |\mathcal{B}_{k,c}|} \sum_{k=1}^K |\mathcal{B}_{k,c}| \mathbf{a}_{k,c}^{(t)} \\
    & = \frac{1}{\sum_{k=1}^K |\mathcal{B}_{k,c}|} \sum_{k=1}^K  \sum_{(\mathbf{x},c) \in \mathcal{B}_{k,c}} f_{\mathrm{mid}}(\mathbf{w}^{(t)},\mathbf{x}) \in \mathbb{R}^d.
\end{aligned}
\end{equation}
We see that each global anchor $\mathbf{a}_c^{(t)} \in \mathbb{R}^d$ still matches with the optimized results in \eqref{eq:global_anchors_compute} and is the midpoint of features of all data that belongs to $\mathcal{B}_{k,c}$, $k \in \{1,2,...,K\}$. All these global anchors $\mathcal{A}^{(t)}=\{ \mathbf{a}_c^{(t)} \}_{c=1}^C$ are then broadcast to be shared by all clients. At this point, the acquired shared anchors are representative of the whole dataset. These are then used to guide the feature learning at each client's local model training.

\paragraph{Step 2: Model Updating} Here we implement the part of the update in~\eqref{eq:global_model_update} in a federated fashion, which generally requires more than one step of SGD update for each round. This involves two substeps: local model training for updating local models, working on the client side, and global model aggregation for aggregating local models from clients, working on the server side.

\emph{Local model training.} Each client conducts local model training on its private dataset with the task supervision and anchor-based feature matching loss. The $k$th local client's model parameter at the communication round $t$ with iteration $r$ is updated as

\begin{equation}
\label{eq:local_model_update}
\mathbf{w}_k^{(t, r+1)} = \mathbf{w}_k^{(t, r)} - \eta \big(  \frac{\partial F_k(\mathbf{w})}{\partial \mathbf{w}} + \lambda \frac{\partial  Q_k(\mathbf{w};\mathcal{A}^{t}) }{\partial \mathbf{w}} \big),
\end{equation}
where $F_k(\cdot)$ and $Q_k(\cdot)$ are the $k$th client's task-specific loss and anchor-based feature matching loss~\eqref{eq:global_objective}, respectively. After $\tau$ iterations of local training, each client $k$ obtains a local model with parameters $\mathbf{w}_k^{(t,\tau)}$ and uploads it to the server.

\emph{Global model aggregation.} The server receives and aggregates the local models from all the clients and obtains an updated global model for the next round; that is, the global model for the next round $t+1$ is obtained as:
\begin{equation}
\label{eq:modelaggr}
    \mathbf{w}^{(t+1)} \leftarrow \sum_{k=1}^K p_k \mathbf{w}_k^{(t,\tau)},
\end{equation}
where $p_k$ is the predefined aggregation weight.

\subsubsection{Strengths of Feature Matching} The proposed anchor-based feature matching can significantly relieve the inconsistency phenomenon in Section~\ref{ssct:me}. Since each client's features are trained to match the shared global anchors, the discrepancy of learned feature spaces across clients can be reduced. Therefore, clients' models establish a more consistent feature space of every category. Furthermore, global anchors contain overall information since they are obtained by aggregating all local anchors. With global anchors, information of other clients is infused into each client, which provides additional guidance on local feature learning especially those categories with relatively few data samples.

\subsection{Contrastive-Guiding Loss}
\label{ssct:cg}
\subsubsection{Method}
To address the problem of overlapping feature space across categories in Section~\ref{ssct:me}, we further propose contrastive-guiding (CG) loss to replace the $\ell_2$-based loss in the feature matching term~\eqref{eq:l2_term}. The idea is to force each feature to be close to the corresponding anchor while keeping far away from non-corresponding anchors. Let $\mathcal{A}^{(t)} = \{ \mathbf{a}_n^{(t)} \}_{n=1}^C$ be the global anchors and 
$\mathbf{w}^{(t, r)}_k$ be the local model for client $k$ at training round $t$ and iteration $r$.  For data sample $\mathbf{x}$, the feature matching loss is
\begin{equation*}
    q \left( f_{\mathrm{mid}}(\mathbf{w}^{(t, r)}_k,\mathbf{x}),  \mathcal{A}^{(t)} | c \right) \ = \  \mathcal{L}_{CE}( \mathbf{s},c),
\end{equation*}
where $\mathcal{L}_{CE}$ is the cross-entropy loss function and $\mathbf{s}=[s_{1}, s_{2}, ..., s_{C}] \in \mathcal{A}^C$ is a similarity vector, whose $n$th element measures the distance with the $n$th anchor:
\begin{equation*}
    s_{n}=\frac{ \exp(  \langle \mathbf{a}_n^{(t)}, f_{\mathrm{mid}}(\mathbf{w}^{(t, r)}_k,\mathbf{x})/\alpha  \rangle ) }{\sum_{i=1}^C \exp( \langle \mathbf{a}_i^{(t)}, f_{\mathrm{mid}}(\mathbf{w}^{(t, r)}_k,\mathbf{x})/\alpha  \rangle )},
\end{equation*}
with a temperature value $\alpha$ determining the level of concentration and $\langle \cdot, \cdot \rangle$ is inner product. By minimizing the cross entropy, we both maximize the similarity between the feature and its corresponding anchor $\mathbf{a}_c^{(t)}$ and minimize the similarity between this feature and each non-corresponding anchor in $\{ \mathbf{a}_n^{(t)} \}_{n \neq c}$.

\begin{figure}[t]
\centering
\includegraphics[scale=0.59]{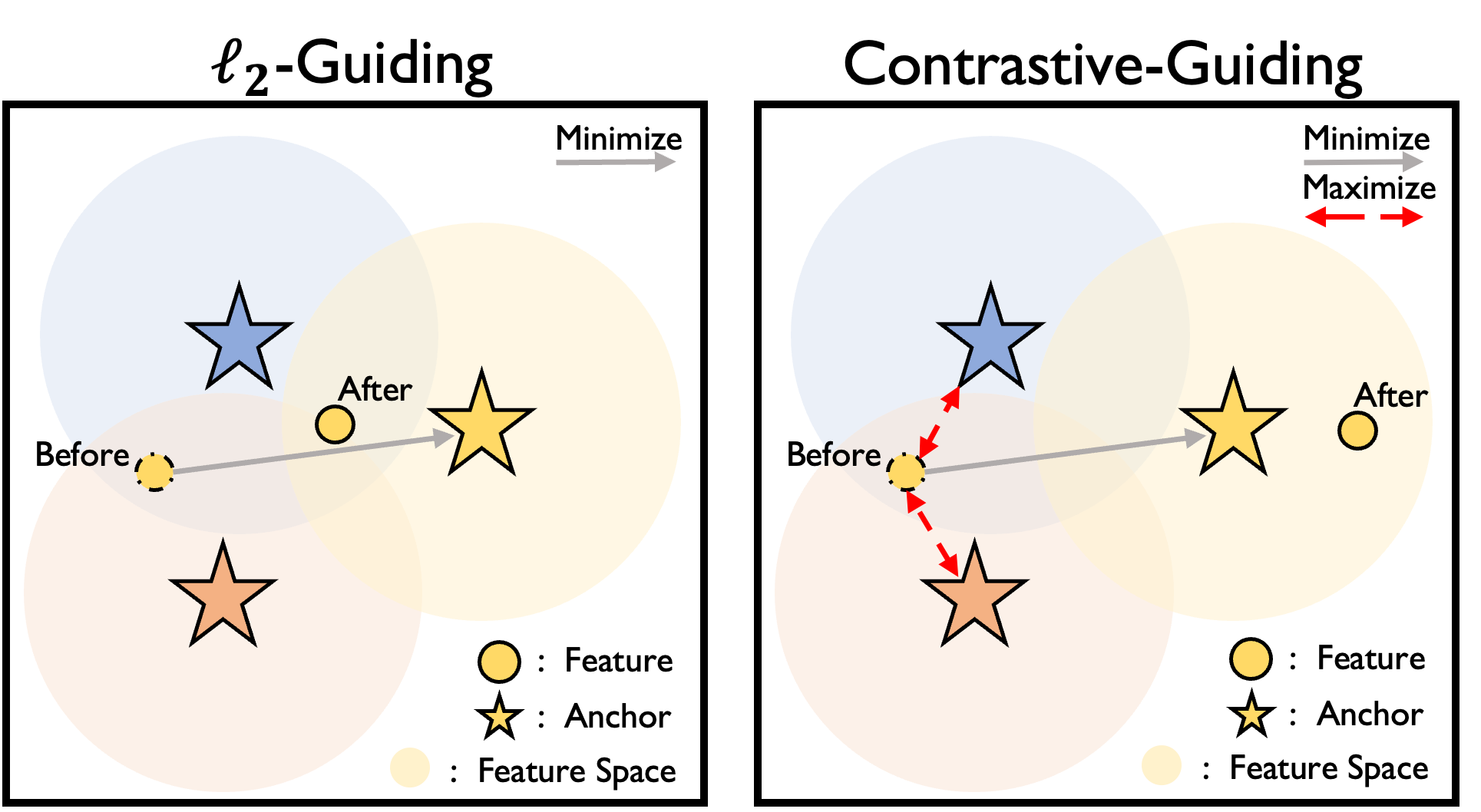}
\caption{Illustration of $\ell_2$-Guiding and contrastive-guiding. Each color denotes one category. $\ell_2$-Guiding only minimizes the distance between the feature and corresponding anchor, which ends up locating the feature at the feature space overlap of categories. However, contrastive-guiding also maximizes the distance between the feature and non-corresponding anchors, which feasibly locates the feature at space that merely belongs to the corresponding category.}
\label{fig:CG}
\end{figure}

\subsubsection{Strengths of CG}
Fig.~\ref{fig:CG} compares the aforementioned $\ell_2$-Guiding and CG. We see that the CG loss achieves more effective feature matching and therefore better targets the overlapping phenomenon from the following two perspectives.

\textbf{i)} CG can provide a more precise target. As shown in the figure, there could be overlap among feature spaces of different categories. For $\ell_2$ loss, only minimizing the distance between the feature and its corresponding anchor could end up locating the feature at that feature space overlap of several categories. However, CG provides a more precise target by simultaneously minimizing the distance between the feature and corresponding anchor and maximizing the distance between the feature and non-corresponding anchors, which feasibly locates the feature at space that merely belongs to the corresponding category.

\textbf{ii)} CG can enlarge the gap across categories. In each round, each feature is further pushed away from non-corresponding anchors, that is, more features are pushed to the non-overlap area as shown in Fig.~\ref{fig:CG}. After this, each anchor (category-wise feature midpoint) is recalculated. Since more features are pushed to the non-overlap area, each recalculated anchor also moves towards the non-overlap area, which ends up enlarging the distance between anchors. This process repeats and eventually the feature space of different categories would be enlarged distinctly, which alleviates the overlap phenomenon.

\subsection{Further Discussions}
\label{ssct:communication_cost}
\subsubsection{Communication Cost} The FedFM method involves two streams of communication, model parameters communication, which is required for most FL methods, and anchors communication, which is relatively negligible. Here, we take a normal setting as example, where the model is ResNet18~\cite{resnet}, category number $C=10$, feature dimension $d=512$. In this case, the anchors cost of each client is $C \times d = 5.12 \times {10}^3$ units while the model cost of each client is $1.17 \times {10}^7$ units. As a result, communicating anchors only requires approximate $0.04\%$ more bandwidth cost.

\subsubsection{Privacy Analysis}
\label{sssec:privacy}
While some feature inversion method~\cite{feature_inversion} attempts to reconstruct image from a single feature, the communicated anchors are the average of a number of features, which makes the reconstruction difficult. This is also verified by~\cite{nofear}, making FedFM a privacy-preserving method. Meanwhile, Secure Aggregation~\cite{bonawitz2017practical} is often used in practice to ensure the safety of model parameters, which can also be adopted to further secure the anchors communication. As CCVR~\cite{nofear} and FedFTG~\cite{fedftg}, the previously illustrated process requires uploading clients' category distributions for weighted aggregating local anchors. For cases when this is prohibited, we can directly compute the simple arithmetic mean, where clients are not asked to upload category distributions. We empirically verify that simple arithmetic mean of anchors achieves comparable performance in Section~\ref{sssct:aggregation_manner}.

\section{Convergence Analysis}
\label{sec:convergence_analysis}

This section provides theoretical convergence analysis of FedFM, including the required assumptions, lemmas and the derived theorem and corollary.

We provide convergence analysis of the global objective function $\Phi(\mathbf{w};\mathcal{A})$ in \eqref{eq:global_objective}, which relies on the following 4 assumptions. In Assumption 1, the assumption of smoothness of $F_k(\mathbf{w})$ is used in standard analysis of SGD and assumptions on $Q_k(\mathbf{w};\mathcal{A})$ are additionally made since that we care about the property of $\Phi_k(\mathbf{w};\mathcal{A})$. Assumptions 2, 3 and 4 are commonly used in the FL literature~\cite{fednova,fedprox,scaffold,convergence_noniid,reddi2020adaptive}. Here, $\ell _2$ loss is applied for simplicity.

\textbf{Assumption 1} (Smoothness). \emph{Each loss function $F_k(\mathbf{w})$ is Lipschitz-smooth. Feature function $f_{\mathrm{mid}}(\mathbf{w}_k,\mathbf{x})$ is Lipschitz-continuous and Lipschitz-smooth.}

\textbf{Assumption 2} (Bounded Scalar). \emph{$\Phi_k(\mathbf{w};\mathcal{A})$ is bounded below by $\Phi_{inf}$.}

\textbf{Assumption 3} (Unbiased Gradient and Bounded Variance). \emph{For each client, the stochastic gradient is unbiased: $\mathbb{E}_\xi[g_k(\mathbf{w}|\xi)] = \nabla \Phi_k(\mathbf{w};\mathcal{A})$, and has bounded variance: $\mathbb{E}_\xi[||g_k(\mathbf{w}|\xi)-\nabla \Phi_k(\mathbf{w};\mathcal{A})||^2] \leq \sigma^2$.}

\textbf{Assumption 4} (Bounded Dissimilarity). \emph{For any set of weights $\{ p_k \geq 0 \}_{k=1}^K$ subject to $\sum_{k=1}^Kp_k=1$, there exists constants $\beta^2 \geq 1$ and $\kappa^2 \geq 0$ such that $\sum_{k=1}^K p_k||\nabla \Phi_k(\mathbf{w};\mathcal{A})||^2 \leq \beta^2 ||\nabla \Phi(\mathbf{w};\mathcal{A})||^2 + \kappa^2$.}

The smoothness property of $\Phi_k(\mathbf{w};\mathcal{A})$ is necessary for convergence analysis. Since $\mathcal{A}$ changes over communication round $t$, smoothness assumption on $Q_k(\mathbf{w};\mathcal{A})$ would be too strong, which requires $T$ assumptions. Thus, we only make one minor assumption on the feature function $f_{\mathrm{mid}}(\mathbf{w}_k,\mathbf{x})$ in Assumption 1 and prove the smoothness of $\Phi_k(\mathbf{w};\mathcal{A})$ as stated in Lemma 1.

\textbf{Lemma 1:}
\emph{The local objective function $\Phi_k(\mathbf{w};\mathcal{A})$ is Lipschitz-smooth: $||\nabla \Phi_k(\mathbf{x};\mathcal{A})- \nabla \Phi_k(\mathbf{y};\mathcal{A})|| \leq L ||\mathbf{x}-\mathbf{y}||$ for some L.}

The fact that $\mathcal{A}$ changes over round $t$ makes it challenging to prove convergence as it changes the global loss function at each round $t$. In Lemma 2, we show that at each communication point, the aggregation and updating of anchors reduces (or keeps) the global loss value.

\textbf{Lemma 2:}
\emph{The global loss function is non-increasing when updating global anchors. That is: $\Phi(\mathbf{w}^{(t+1)};\mathcal{A}^{(t+1)}) \leq \Phi(\mathbf{w}^{(t+1)};\mathcal{A}^{(t)})$.}

Based on this key lemma, we derive our main Theorem, which is stated as follows:

\textbf{Theorem 1 (Optimization bound of the global objective function).} \emph{Under Assumptions 1 to 4, if we set $\eta L \leq min\{ \frac{1}{2 \tau}, \frac{1}{\sqrt{2 \tau (\tau-1)(2 \beta^2 +1)}} \}$, the optimization error will be bounded as follows:}
\begin{equation*}
\begin{aligned}
& \mathop{\min}_{t} \mathbb{E} ||\nabla \Phi (\mathbf{w}^{(t)};\mathcal{A}^{(t)})||^2 \\
\leq & \frac{4[\Phi(\mathbf{w}^{(0)};\mathcal{A}^{(0)})-\Phi_{inf}]}{\tau \eta T}+4\eta L \sigma^2 \sum_{k=1}^K p_k^2\\
    & + 3 (\tau-1) \eta^2 \sigma^2L^2 + 6\tau (\tau-1) \eta^2 L^2 \kappa^2,
\end{aligned}
\end{equation*}
\emph{where $\eta$ is the client learning rate and $\tau$ is the number of local iterations.}

Theorem 1 indicates that as $T \rightarrow \infty$, the expectation of optimization error will be bounded by a constant number for fixed $\eta$. When setting a proper learning rate $\eta$, we have the following corollary:

\textbf{Corollary 1 (Convergence of the global objective function)} \emph{By setting $\eta = \frac{1}{\sqrt{\tau T}}$, FedFM will converge to a stationary point. Specifically, the bound could be further optimized as:}
\begin{equation*}
\begin{aligned}
& \mathop{\min}_{t} \mathbb{E} ||\nabla \Phi (\mathbf{w}^{(t)};\mathcal{A}^{(t)})||^2 \\
\leq & \frac{4[\Phi(\mathbf{w}^{(0)};\mathcal{A}^{(0)})-\Phi_{inf}]+4 L \sigma^2 \sum_{k=1}^K p_k^2}{\sqrt{\tau T}}\\
    & + \frac{3 (\tau-1) \sigma^2L^2}{\tau T} + \frac{6\tau (\tau-1) L^2 \kappa^2}{\tau T}\\
= & \mathcal{O}(\frac{1}{\sqrt{\tau T}}) + \mathcal{O}(\frac{1}{T}) + \mathcal{O}(\frac{\tau}{T}).
\end{aligned}
\end{equation*}

This corollary indicates that as $T \rightarrow \infty$, the error's upper bound approaches $0$. Also, given a finite $T$, there exists a best $\tau$ that minimizes the error's upper bound. These analyses show that FedFM can achieve the same convergence rate as most methods, such as FedAvg~\cite{convergence_noniid,fednova}. Therefore, FedFM can achieve feature matching across clients without affecting its convergence. The detailed proof is included in the Appendix.

\section{An Efficient Variant: FedFM-Lite}
\label{sec:fedfm_lite}

\begin{figure}[t]
\centering
\includegraphics[scale=0.5]{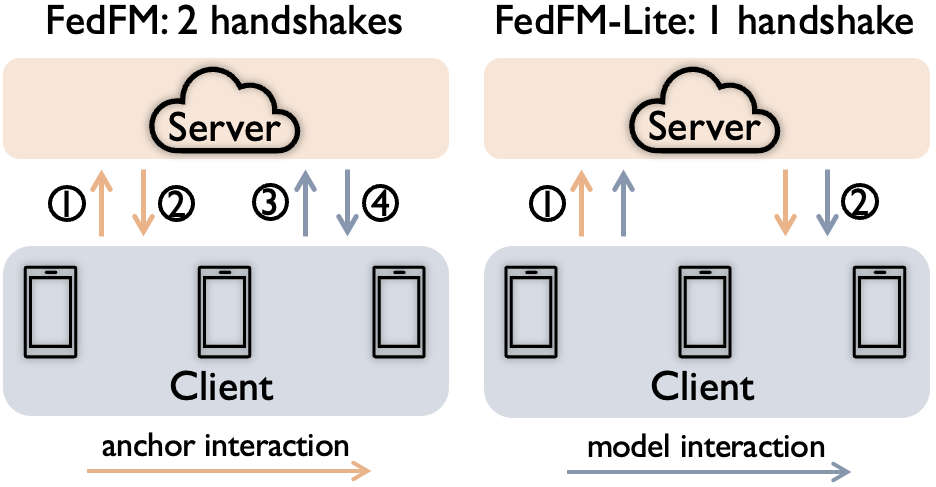}
\caption{Comparison between FedFM and FedFM-Lite. FedFM introduces minor bandwidth cost but requires 2 handshakes each round. FedFM-Lite further eliminates this issue, which introduces minor bandwidth cost and requires only 1 handshake.}
\label{fig:overview_fedfm_lite}
\end{figure}

We also propose an efficient and flexible variant of FedFM, called FedFM-Lite. The previously proposed FedFM involves two separate communication flows, anchors communication before local model training and models communication after local model training. Though we have discussed that the anchors communication introduces minor bandwidth cost, FedFM requires twice handshakes between client and server within a federated round, which could be a drawback in real-world application since that each handshake requires some synchronization time. Thus, to mitigate this issue, we propose a more efficient variant, FedFM-Lite, which requires one handshake between client and server. We compare FedFM and FedFM-Lite in Fig.~\ref{fig:overview_fedfm_lite}.

In FedFM-Lite, each client computes local anchors after the process of local model training and sends the local anchors together with model parameters. In this way, models and anchors are communicated within one handshake, which saves some synchronization time and makes it more efficient in real-world application. Beside higher efficiency, FedFM-Lite is also more flexible for real-world implementation since that models and anchors can be communicated at different frequencies. As discussed before, communicating anchors requires much less costs compared with model parameters. This motivates us to consider reducing the frequency of models communication and utilize anchors communication as compensation. Specifically, in a federated learning process that consists of $T$ communication rounds, we can communicate anchors for each round while communicate models for every $a$ rounds. This results in roughly $a$ times less communication costs compared with most existing methods, such as FedAvg~\cite{fedavg}, FedProx~\cite{fedprox}, FedDyn~\cite{feddyn} and $2 \times a$ less than SCAFFOLD~\cite{scaffold}.

\section{Experiments}
\label{sec:experiments}

This section presents experimental details and results. We compare our proposed FedFM with state-of-the-art methods on various heterogeneous settings and datasets, which are evaluated by accuracy, feature space quality, memory and communication bandwidth cost.

\subsection{Experimental Setup}
\paragraph{Federated Setting} We set the number of clients $K=10$ and conduct experiments on datasets including CIFAR-10~\cite{cifar10}, CINIC-10~\cite{darlow2018cinic} and CIFAR-100. Here we consider three data heterogeneity (non-IID) settings. 1) NIID-1: the category distributions of clients follow a Dirichlet distribution $Dir_{10}(\beta)$, where $\beta$ (default $0.5$) correlates to the heterogeneity level, which is a widely considered setting~\cite{fedma,bayesian}; 2) NIID-2: each client has several dominant categories (with much more data samples) while we keep the dataset size of each client the same. We consider this setting to focus on the distribution heterogeneity but not quantity imbalance; 3) NIID-3: each client has no data sample from several categories, which is also considered in~\cite{fedavg,fedprox}. Fig.9 in appendix shows the data distribution of these three Non-IID settings on CIFAR-10.

\paragraph{Implementation} We run $T=100$ communication rounds for all experiments. In each round, every client runs for $10$ local epochs with a batch size of $64$. We apply ResNet18~\cite{resnet} for CIFAR-10~\cite{cifar10} and CINIC-10~\cite{darlow2018cinic}, ResNet50 for CIFAR-100~\cite{cifar10}. We use SGD optimizer with learning rate $0.01$, weight decay rate $1e^{-5}$ and SGD momentum $0.9$. These are commonly used experimental settings~\cite{moon,nofear}. For evaluation, we hold out a testing dataset at the server side and conduct the above non-IID partitions on the training set. For each client, $20\%$ of the training set is held out for validation. We average the results on each local validation set and save the best model. Finally, we report the testing accuracy of the best model on the testing dataset.

We consider $f_{\mathrm{full}}(\cdot,\cdot)$ as a standard ResNet and the feature extractor $f_{\mathrm{mid}}(\cdot,\cdot)$ as $f_{\mathrm{full}}(\cdot,\cdot)$ without the last fully-connected layer. For feature matching, the feature is normalized before applying the feature matching loss term. FedFM denotes FedFM with CG unless explicitly specified. We run FedAvg for the first $T_s$ rounds and then launch our proposed FedFM. For all methods, we tune the hyper-parameters in a reasonable range and report the best results. Generally, for FedFM, $\lambda=50.0$ and $T_s=20$ is a relatively better choice.

\begin{table*}[t]
\begin{center}
\caption{Classification accuracy ($\%$) under NIID-1 and NIID-2 settings on CIFAR-10~\cite{cifar10}, CINIC-10~\cite{darlow2018cinic} and CIFAR-100. NIID-1 is under Dirichlet distribution $Dir_{10}(0.5)$ and NIID-2 is the distribution where each client has one dominant category. Memory shows the required number of floating numbers of each client in each round ($\times {10}^3$). FedFM consistently outperforms other state-of-the-art methods with relatively less memory cost across various settings.}
\label{table:main}
\begin{tabular}{c|ccc|ccc|ccc}
\hline
\multirow{2}{*}{Method} & \multicolumn{3}{c|}{CIFAR-10} & \multicolumn{3}{c|}{CINIC-10} & \multicolumn{3}{c}{CIFAR-100} \\ \cline{2-10} 
        & NIID-1 & NIID-2 & Memory & NIID-1 & NIID-2 & Memory & NIID-1 & NIID-2 & Memory \\ \hline
FedAvg~\cite{fedavg}     & 66.69 $\pm 0.69$ & 69.47 $\pm 0.48$ & 11,182 & 55.96 $\pm 0.16$ & 58.56 $\pm 0.22$ & 11,182 & 62.16 $\pm 0.04$ & 62.33 $\pm 0.27$ & 23,705 \\
FedAvgM~\cite{fedavgm}   & 66.85 $\pm 0.42$ & 67.87 $\pm 0.17$ & 11,182 & 56.15 $\pm 0.45$ & 58.79 $\pm 0.30$ & 11,182 & 61.23 $\pm 0.12$ & 61.30 $\pm 0.27$ & 23,705 \\
FedProx~\cite{fedprox}   & 66.99 $\pm 0.26$ & 69.42 $\pm 0.38$ & 22,364 & 55.58 $\pm 0.13$ & 58.32 $\pm 0.11$ & 22,364 & 61.96 $\pm 0.05$ & 62.20 $\pm 0.28$ & 47,410\\
SCAFFOLD~\cite{scaffold} & 69.91 $\pm 0.54$ & 71.48 $\pm 0.23$ & 22,364 & 58.60 $\pm 0.27$ & 60.78 $\pm 0.32$ & 22,364 & 67.32 $\pm 0.29$ & 67.24 $\pm 0.03$ & 47,410\\
FedDyn~\cite{feddyn}     & 68.32 $\pm 0.34$ & 67.63 $\pm 0.16$ & 22,364 & 56.71 $\pm 0.50$ & 59.92 $\pm 0.15$ & 22,364 & 43.41 $\pm 0.54$ & 46.44 $\pm 0.87$ & 47,410\\
FedNova~\cite{fednova}   & 66.80 $\pm 0.81$ & 69.45 $\pm 0.49$ & 11,182 & 55.67 $\pm 0.24$ & 58.63 $\pm 0.22$ & 11,182 & 62.35 $\pm 0.20$ & 62.31 $\pm 0.26$ & 23,705 \\
MOON~\cite{moon}         & 67.74 $\pm 0.30$ & 71.09 $\pm 0.22$ & 33,546 & 57.25 $\pm 0.07$ & 59.28 $\pm 0.03$ & 33,546 & 62.56 $\pm 0.22$ & 62.99 $\pm 0.13$ & 71,115\\
\textbf{FedFM (ours)}    & \textbf{72.89} $\pm 0.22$ & \textbf{74.52} $\pm 0.21$ & 11,187 & \textbf{62.56} $\pm 0.40$ & \textbf{65.75} $\pm 0.46$ & 11,187 &  \textbf{71.48} $\pm 0.25$ & \textbf{72.13} $\pm 0.45$ & 23,909 \\ \hline
\end{tabular}
\end{center}
\end{table*}

\begin{table*}[t]
\begin{center}
\caption{Classification accuracy ($\%$) under NIID-3 setting on CIFAR-10. Missing $x$ setting represents that each client has no data sample of $x$ categories. Memory and Bandwidth show the required number of floating numbers of each client in each round ($\times {10}^3$). Our proposed FedFM consistently outperforms other state-of-the-art methods with minor additional resource cost.}
\label{table:missing}
\begin{tabular}{cccccccc}
   \hline
   Method & Missing 1 & Missing 2 & Missing 3 & Missing 5 & Missing 7 & Memory & Bandwidth \\
   \hline
   FedAvg~\cite{fedavg}     & 70.54 $\pm 0.22$ & 70.50 $\pm 0.24$ & 69.87 $\pm 0.30$ & 67.25 $\pm 0.54$ & 59.52 $\pm 0.59$ & 11,182 & 11,182 \\
   FedAvgM~\cite{fedavgm}   & 70.02 $\pm 0.40$ & 69.93 $\pm 0.57$ & 69.34 $\pm 0.37$ & 67.04 $\pm 0.47$ & 57.08 $\pm 0.66$ & 11,182 & 11,182 \\
   FedProx~\cite{fedprox}   & 71.16 $\pm 0.42$ & 70.72 $\pm 0.35$ & 69.82 $\pm 0.23$ & 67.25 $\pm 0.54$ & 58.58 $\pm 0.23$ & 22,364 & 11,182 \\
   SCAFFOLD~\cite{scaffold} & 72.67 $\pm 0.39$ & 72.94 $\pm 0.30$ & 72.60 $\pm 0.22$ & 71.43 $\pm 0.05$ & 64.28 $\pm 0.60$ & 22,364 & 22,364 \\
   FedDyn~\cite{feddyn}     & 67.43 $\pm 0.51$ & 67.76 $\pm 0.64$ & 67.78 $\pm 0.28$ & 69.53 $\pm 0.59$ & 64.75 $\pm 0.30$ & 22,364 & 11,182 \\
   FedNova~\cite{fednova}   & 70.56 $\pm 0.25$ & 70.48 $\pm 0.23$ & 70.05 $\pm 0.15$ & 67.56 $\pm 0.52$ & 59.66 $\pm 0.42$ & 11,182 & 11,182 \\
   MOON~\cite{moon}         & 72.64 $\pm 0.25$ & 72.21 $\pm 0.22$ & 71.57 $\pm 0.23$ & 68.86 $\pm 0.27$ & 57.80 $\pm 1.02$ & 33,546 & 11,182 \\
   \textbf{FedFM (ours)}    & \textbf{75.97} $\pm 0.44$ & \textbf{75.84} $\pm 0.23$ & \textbf{75.04} $\pm 0.29$ & \textbf{73.23} $\pm 0.35$ & \textbf{65.24} $\pm 0.52$ & 11,187 & 11,187 \\
   \hline
\end{tabular}
\end{center}
\end{table*}

\subsection{Main Results}

We compare FedFM with seven existing classical methods, including FedAvg~\cite{fedavg}, FedAvgM~\cite{fedavgm}, FedProx~\cite{fedprox}, SCAFFOLD~\cite{scaffold}, FedDyn~\cite{feddyn}, FedNova~\cite{fednova} and MOON~\cite{moon} on various non-IID settings and datasets. We first show accuracy comparisons quantitatively and then demonstrate qualitative comparisons in feature space by T-SNE~\cite{tsne} visualization.

\subsubsection{Quantitative Analysis}

Table \ref{table:main} presents accuracy comparisons on three datasets under both NIID-1 and NIID-2 settings. For each entry in the table, we run three independent trials and report the mean and standard deviation results. We see that i) FedFM consistently outperforms other state-of-the-art methods on all tasks. ii) On the relatively more complicated task, NIID-1 setting on CIFAR100, the proposed FedFM significantly outperforms other methods. Specifically, compared with standard FL, FedAvg~\cite{fedavg}, FedFM achieves $9.40\%$ higher accuracy. iii) On the six different tasks, FedFM outperforms the second-best method (SCAFFOLD~\cite{scaffold}) $2.98\%, 3.04\%, 3.96\%, 4.97\%, 4.16\%, 4.89\%$, respectively. It is also worth mentioning that SCAFFOLD~\cite{scaffold} requires roughly twice the memory and communication bandwidth costs.

Table \ref{table:missing} shows the accuracy, memory, bandwidth comparisons under NIID-3 setting on CIFAR-10. In NIID-3, each client has no data sample from several ($x$) categories, which is denoted as Missing $x$ setting. We conduct experiments on different $x \in \{ 1, 2, 3, 5, 7 \}$.  We see that i) the performances of all methods degrade as $x$ increases since larger $x$ corresponds to a more heterogeneous setting. This verifies that data heterogeneity significantly affects the performance of FL. ii) FedFM consistently outperforms other state-of-the-art methods. Specifically, it outperforms FedAvg~\cite{fedavg} by $5.53\%$ and SCAFFOLD~\cite{scaffold} by $2.28\%$ on average. iii) Compared with FedAvg~\cite{fedavg}, FedFM achieves significantly better performance while introducing minor memory and bandwidth costs. Compared with SCAFFOLD~\cite{scaffold}, FedFM achieves better performance with nearly half of the memory and bandwidth costs.

\begin{figure*}[t]
\centering
\subfloat[FedAvg~\cite{fedavg}]{\includegraphics[width=1.5in]{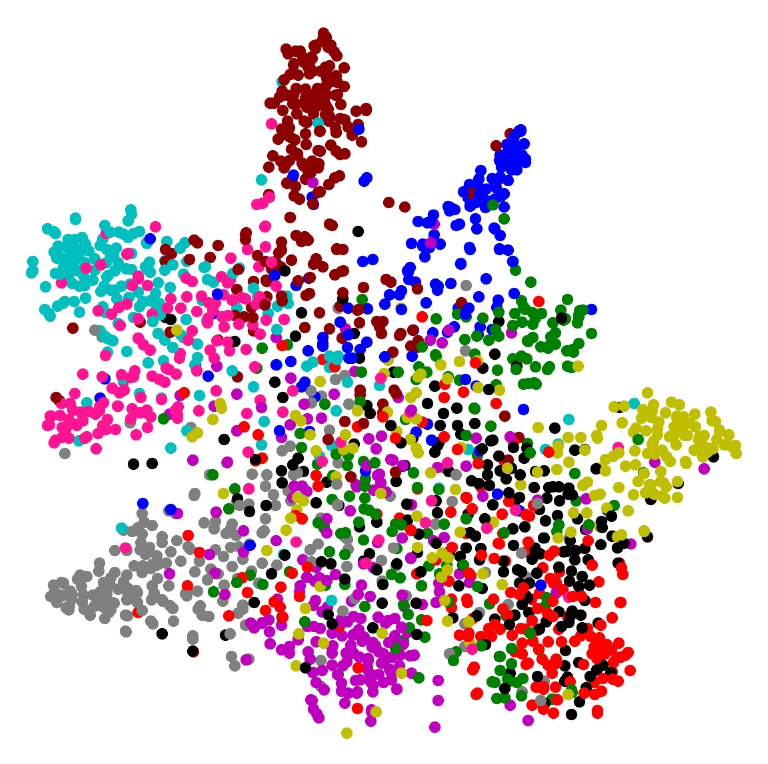}%
}
\hfil
\subfloat[FedAvgM~\cite{fedavgm}]{\includegraphics[width=1.5in]{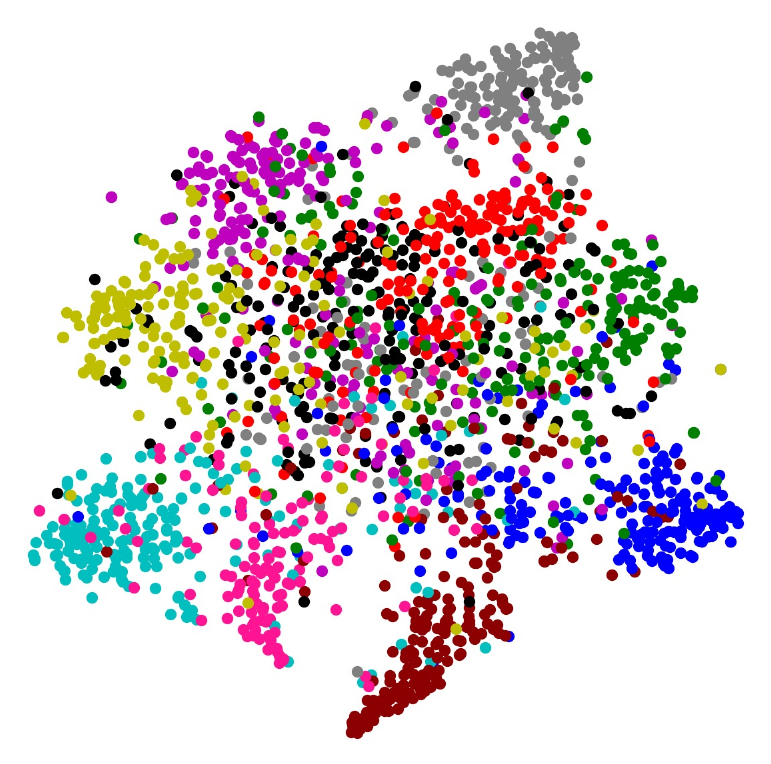}%
}
\hfil
\subfloat[FedProx~\cite{fedprox}]{\includegraphics[width=1.5in]{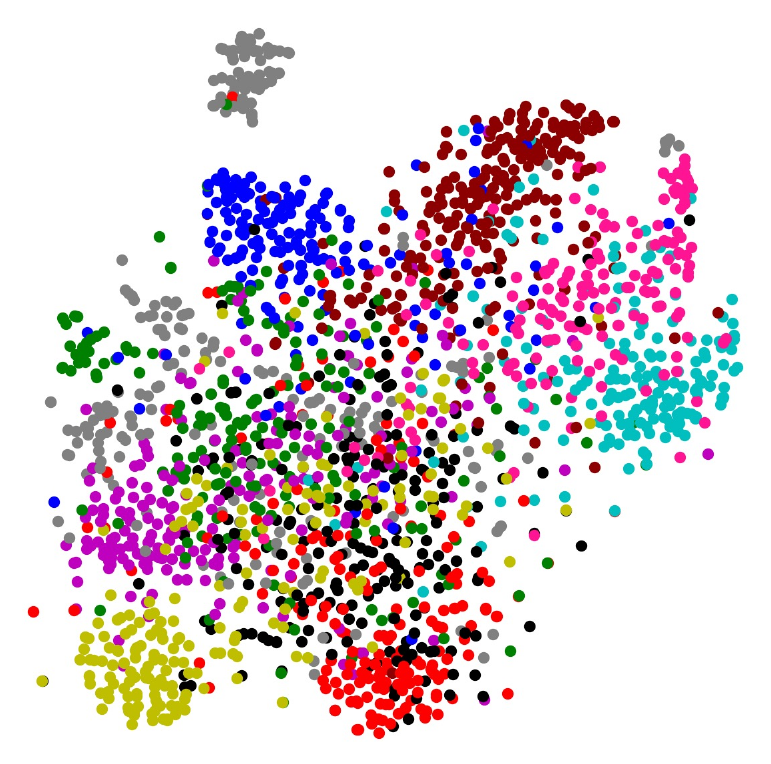}%
}
\hfil
\subfloat[SCAFFOLD~\cite{scaffold}]{\includegraphics[width=1.5in]{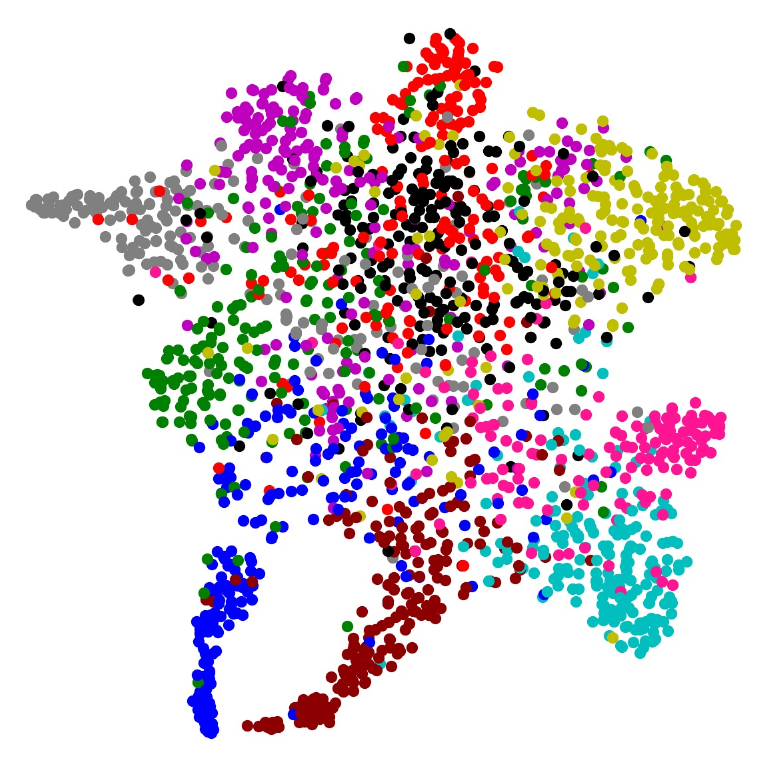}%
}
\hfil
\subfloat[FedDyn~\cite{feddyn}]{\includegraphics[width=1.5in]{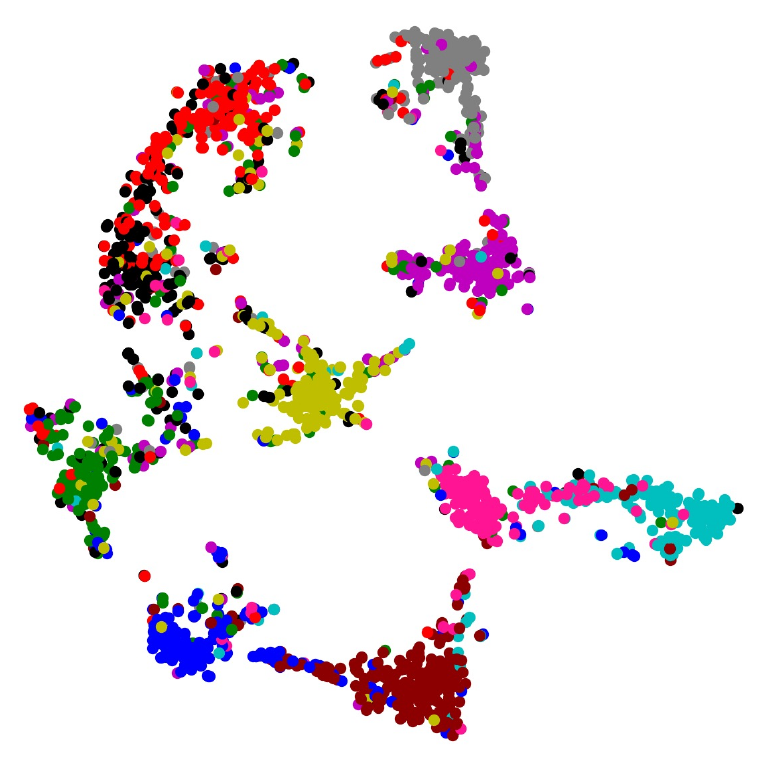}%
}
\hfil
\subfloat[FedNova~\cite{fednova}]{\includegraphics[width=1.5in]{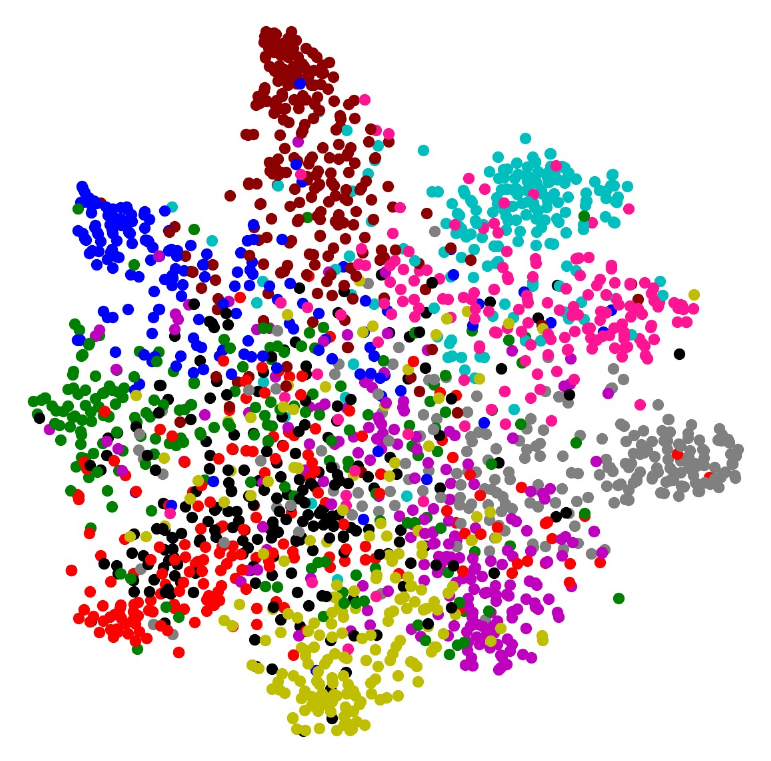}%
}
\hfil
\subfloat[MOON~\cite{moon}]{\includegraphics[width=1.5in]{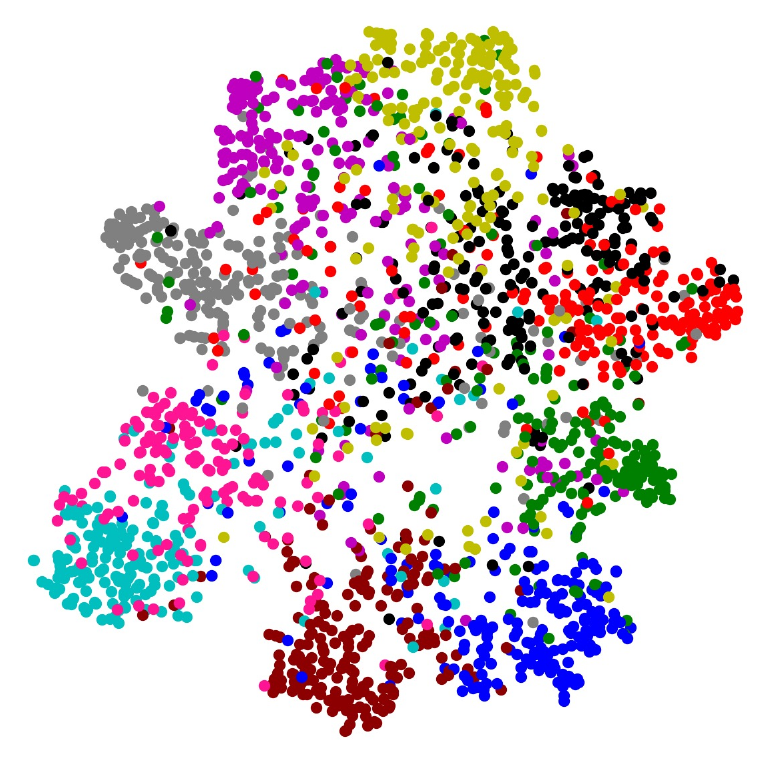}%
}
\hfil
\subfloat[\textbf{FedFM (ours)}]{\includegraphics[width=1.5in]{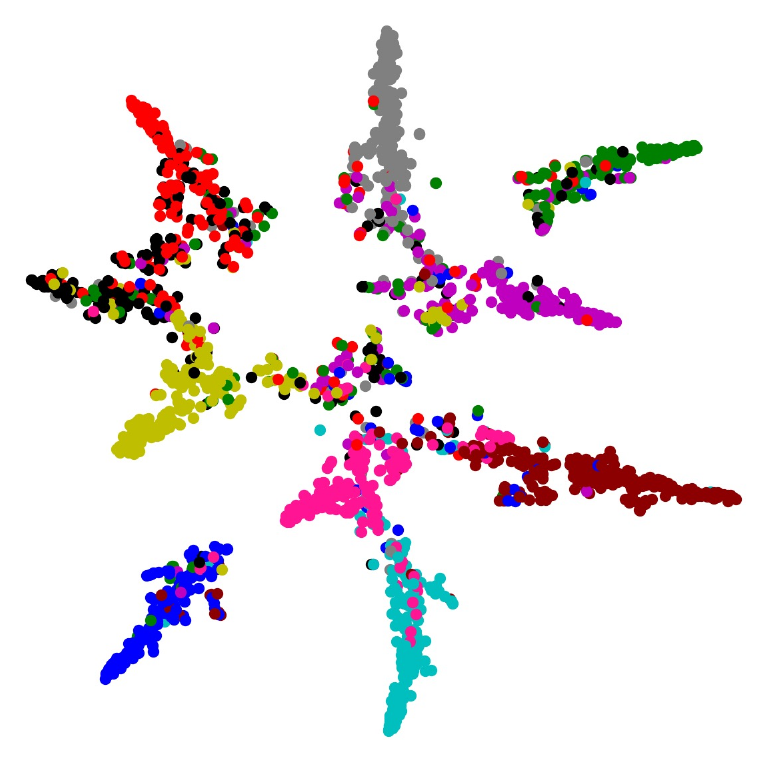}%
}
\hfil
\caption{Qualitative comparisons among methods through T-SNE~\cite{tsne} visualization. Each dot represents the feature of one data sample, whose color denotes its category. FedFM establishes the most compact and distinct clusters in feature space. FedDyn~\cite{feddyn} has moderately good visualization results but only achieves $68.32\%$ accuracy while FedFM achieves $72.89\%$.}
\label{fig:tsne}
\end{figure*}

\begin{table*}[t]
\begin{center}
\caption{Numerical quality evaluation of feature spaces. Higher NMI and SS correspond to higher quality of feature spaces. Our proposed FedFM achieves significantly highest NMI and SS.}
\label{table:quality}
\begin{tabular}{ccccccccc}
   \hline
   Metric & FedAvg~\cite{fedavg} & FedAvgM~\cite{fedavgm} & FedProx~\cite{fedprox} & SCAFFOLD~\cite{scaffold} & FedDyn~\cite{feddyn} & FedNova~\cite{fednova} & MOON~\cite{moon} & \textbf{FedFM (ours)}  \\
   \hline
   NMI & 0.413 & 0.411 & 0.397 & 0.432 & 0.485 & 0.416 & 0.481 & \textbf{0.557} \\
   SS & 0.036 & 0.038 & 0.006 & 0.056 & 0.136 & 0.049 & 0.068 & \textbf{0.173}\\
   \hline
\end{tabular}
\end{center}
\end{table*}

\subsubsection{Qualitative Analysis}
Fig.~\ref{fig:tsne} presents T-SNE~\cite{tsne} visualization results in feature space of each method, where different color denote different category. All the sampled data is fed into the final global model of each method to obtain the corresponding features, which are then plot using T-SNE~\cite{tsne}. We see that i) most previous methods suffer from slack category-wise feature space while our FedFM establishes significantly more compact category-wise feature space, which reflects the effectiveness of utilizing anchors to conduct feature matching. ii) Most previous methods suffer from ambiguous boundaries. However, our FedFM establishes clusters with clear boundaries and large gap which are contributed by using anchors to attract features and the contrastive-guiding loss. These two phenomena indicate that our proposed FedFM indeed benefits the establishment of feature space and gives the evidence of significantly improved performance. Note that though FedDyn~\cite{feddyn} seems to establish a good feature space, it only achieves a $68.32\%$ accuracy, which is significantly lower than that of our proposed FedFM ($72.89\%$).

For more comprehensive comparisons, we also evaluate the quality of feature space in Fig.~\ref{fig:tsne} using normalized mutual information (NMI) and silhouette score (SS)~\cite{rousseeuw1987silhouettes}. NMI is capable of measuring the quality of clustering. SS is a measure of how similar an object is to its own cluster compared to other clusters. Note that for NMI, we first apply the K-Means~\cite{lloyd1982least} to perform clustering on sampled features of all methods and then use NMI to measure the clustering quality. Both NMI and SS are mesured using Scikit-learn~\cite{pedregosa2011scikit}. Higher NMI and SS correspond to higher quality of feature spaces. We present the evaluation results in Table~\ref{table:quality}. The table shows that our proposed FedFM achieves significantly higher NMI and SS. Specifically, compared with FedDyn~\cite{feddyn}, FedFM achieves $14.8\%$ higher NMI and $27.2\%$ higher SS. This gives evidence for the seemly great feature space but ordinary accuracy performance of FedDyn~\cite{feddyn} in a way.

\subsection{Further Comparisons}

\subsubsection{Comparisons with FedProto}

Targeting a different task, personalization in FL, FedProto~\cite{tan2021fedproto} uses prototype to provide feature information from others to enhance personalization while FedFM focuses on generalization in FL. To further verify their difference, we implement a generalized version of FedProto~\cite{tan2021fedproto} and compare it with FedFM under NIID-1 setting on CIFAR-10~\cite{cifar10}. Experiments show that generalized FedProto achieves $67.33 \pm 0.49 \%$ accuracy, which is outperformed by SCAFFOLD~\cite{scaffold}, FedDyn~\cite{feddyn} and MOON~\cite{moon}. Our proposed FedFM achieves $72.89 \pm 0.22 \%$ accuracy, which outperforms generalized FedProto~\cite{tan2021fedproto} by $5.56\%$.

\subsubsection{Performance and Resource Costs}

FedFM introduces minor resource costs while bringing significantly better performance. To verify this point, we conduct experiments on CIFAR-100~\cite{cifar10} under various client numbers ($K \in \{20, 30, 50, 100\}$). Beside accuracy comparisons, we also compare the memory and bandwidth cost of these methods, which are evaluated by the number of required floating numbers ($\times 10^8$) for the overall FL process. We present the results in Table~\ref{table:performance_cost}. Experiments show that i) our proposed FedFM consistently outperforms state-of-the-art methods for various client numbers, indicating its applicability to scenario with large client number. ii) FedFM achieves significantly better performance with minor additional memory and bandwidth overhead. Specifically, FedFM takes only $0.86\%$ more communication overhead to achieve $13.97\%$ better classification performance than FedAvg~\cite{fedavg} when $K=100$. Compared with the second-best method (SCAFFOLD~\cite{scaffold}), FedFM achieves $7.68\%$ higher accuracy when $K=100$ with only half the memory and bandwidth costs.

\begin{table*}[ht]
\begin{center}
\caption{Comparisons of accuracy, memory and bandwidth costs on CIFAR-100. Each entry shows classification accuracy ($\%$). Within parentheses, it shows the required memory / bandwidth cost, evaluated by floating numbers ($\times 10^8$). When $K=100$, i) FedFM takes only \textbf{$0.86\%$} more resource overhead to achieve $13.97\%$ higher accuracy than FedAvg~\cite{fedavg}, ii) FedFM achieves $7.68\%$ higher accuracy with only half the memory and bandwidth costs compared with SCAFFOLD~\cite{scaffold}.}
\label{table:performance_cost}
\begin{tabular}{cccccc}
   \hline
   K & 20 & 30 & 50 & 100  \\
   \hline
   FedAvg~\cite{fedavg}     & 58.48 (474 / 474) & 54.46 (711 / 711) & 50.20 (1,185 / 1,185) & 41.41 (2,370 / 2,370) \\
   FedAvgM~\cite{fedavgm}   & 58.36 (474 / 474) & 54.48 (711 / 711) & 52.86 (1,185 / 1,185) & 46.72 (2,370 / 2,370) \\
   FedProx~\cite{fedprox}   & 58.27 (948 / 474) & 54.50 (1,422 / 711) & 50.55 (2,370 / 1,185) & 40.62 (4,740 / 2,370)\\
   SCAFFOLD~\cite{scaffold} & 64.65 (948 / 948) & 61.82 (1,422 / 1,422) & 56.71 (2,370 / 2,370) & 47.70 (4,740 / 4,740) \\
   FedDyn~\cite{feddyn}     & 41.90 (948 / 474) & 41.13 (1,422 / 711) & 39.30 (2,370 / 1,185) & 31.21 (4,740 / 2,370) \\
   FedNova~\cite{fednova}   & 58.01 (474 / 474) & 53.83 (711 / 711) &50.34 (1,185 / 1,185) & 42.61 (2,370 / 2,370)\\
   MOON~\cite{moon}         & 57.63 (1,422 / 474) & 52.71 (2,133 / 711) & 47.84 (3,555 / 1,185)& 38.45 (7,110 / 2,370)\\
   \textbf{FedFM (ours)} & \textbf{69.49} (478 / 478) & \textbf{67.70} (717 / 717) & \textbf{64.22} (1,195 / 1,195) & \textbf{55.38} (2,390 / 2,390)\\
   \hline
\end{tabular}
\end{center}
\end{table*}

\begin{figure}[t]
\centering
\includegraphics[scale=0.4]{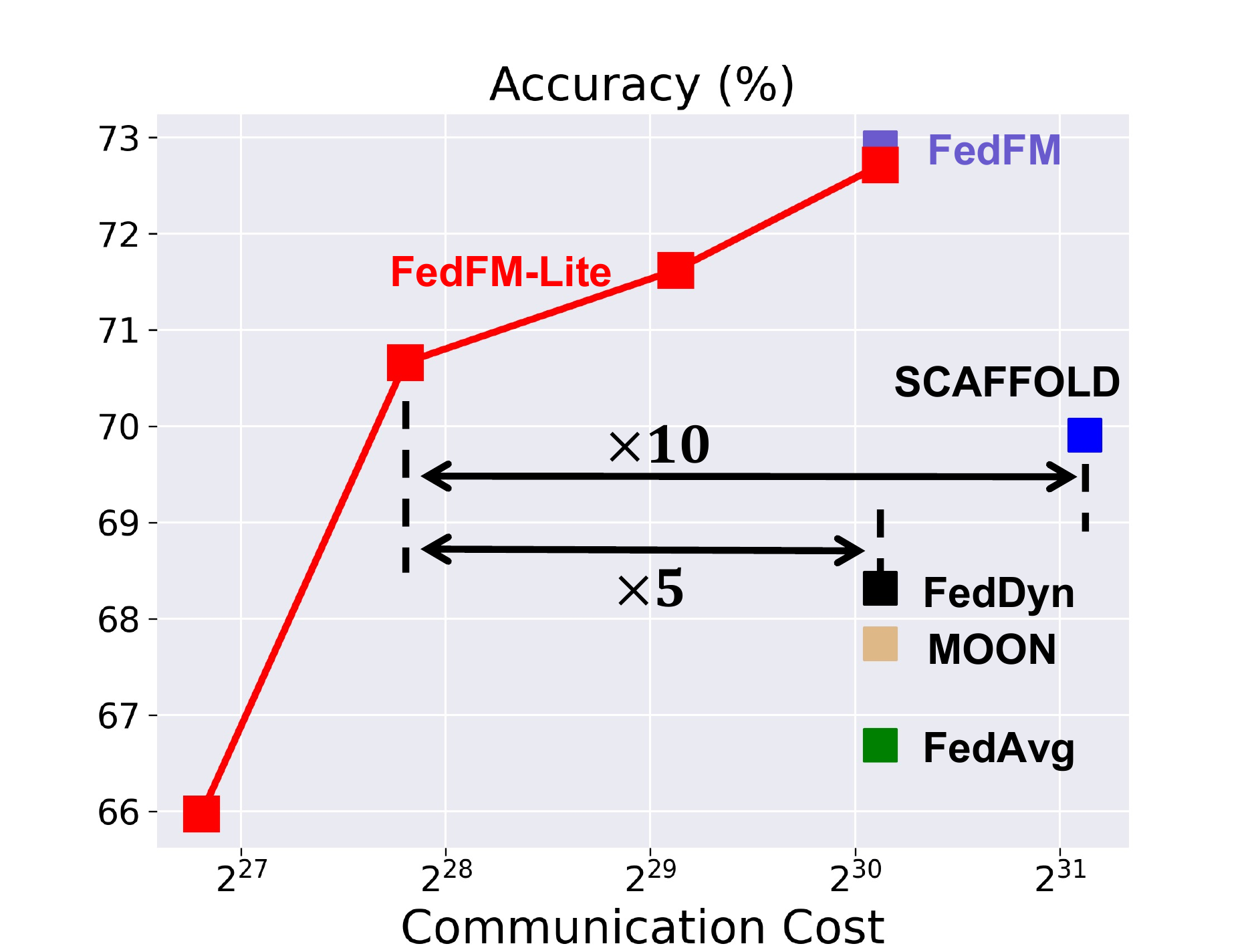}
\caption{Comparison of performances and communication costs among FedFM-Lite and several classical methods. Our proposed FedFM-Lite can achieve significantly better performance while saving 5 or even 10 times communication costs compared with existing techniques.}
\label{fig:exp_fedfm_lite}
\end{figure}

\subsubsection{Performance of An Efficient Variant: FedFM-Lite}
    
In Section~\ref{sec:fedfm_lite}, we propose an efficient variant, called FedFM-Lite, which is more efficient and flexible in practice. Since anchors require less communication bandwidth costs than model communication, we propose to reduce the communication frequency of model communication while keeping the communication frequency of anchor communication, which can be easily achieved in FedFM-Lite. This modification saves communication bandwidth costs to a large extent.

To empirically verify this efficiency and flexibility, we conduct the following experiments. We communicate anchors for each round while we communicate models every $a \in \{1, 2, 5, 10\}$ round(s), that is, larger $a$ corresponds to less communication cost. We show the communication cost and final performance of each trial in Fig.~\ref{fig:exp_fedfm_lite}. We also present several representative methods for comparison.

Experiments show that when we communicate models every $a \in \{1, 2, 5\}$ communication round(s), FedFM-Lite can significantly outperform compared methods. Specifically, when $a=5$, FedFM-Lite outperforms FedDyn~\cite{feddyn} by $2.34\%$ with $5$ times less communication costs and SCAFFOLD~\cite{scaffold} by $0.75\%$ with $10$ times less communication costs. These experiments show that for bandwidth limited scenarios, FedFM-Lite can be an efficient candidate algorithm.

\begin{table}[t]
\begin{center}
\caption{Effects of global anchors aggregation manner. (Weighted) denotes aggregating category-wise anchors according to each client's number of corresponding samples. (Uniform) denotes computing simple arithmetic mean of anchors. Experiments show that FedFM (Uniform) performs comparably with FedFM (Weighted) while FedFM (Uniform) does not need uploading client's category distribution.}
\label{table:anchor_aggregation}
\begin{tabular}{cccc}
   \hline
   Method & FedAvg~\cite{fedavg} & FedFM (Weighted) & FedFM (Uniform) \\ \hline
   Accuracy & 66.69 $\pm 0.69$ & 72.89 $\pm 0.22$ & 72.87 $\pm 0.26$\\
   \hline
\end{tabular}
\end{center}
\end{table}

\subsection{Ablation Study}

\begin{figure*}[!t]
\centering
\subfloat[Modularity of FM]{\includegraphics[width=2.1in]{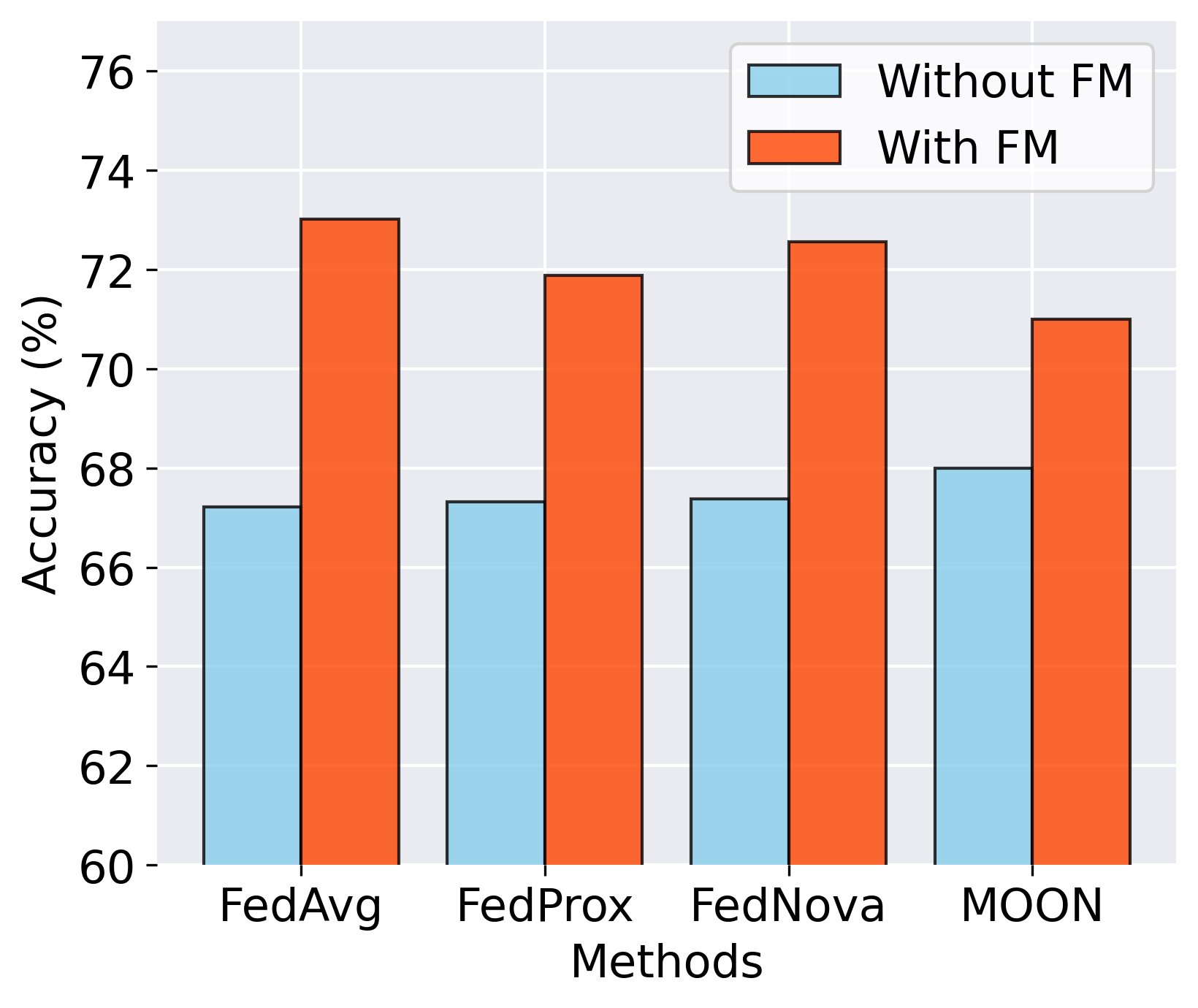}%
\label{fig:abl_modularity}}
\hfil
\subfloat[$\ell_2$-Guiding and CG]{\includegraphics[width=2.1in]{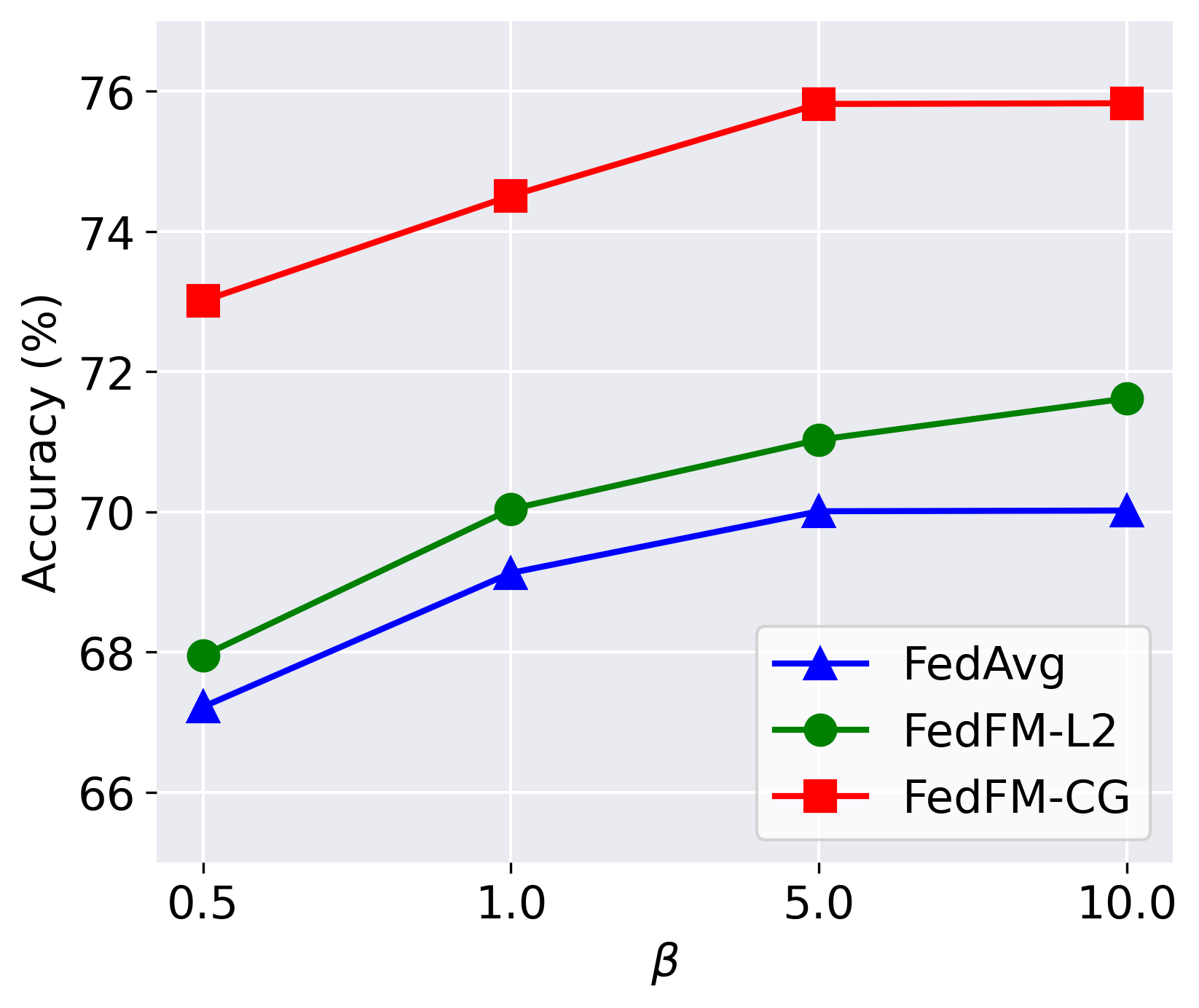}%
\label{fig:abl_cg}}
\hfil
\subfloat[$T_s$ and $\lambda$]{\includegraphics[width=2.1in]{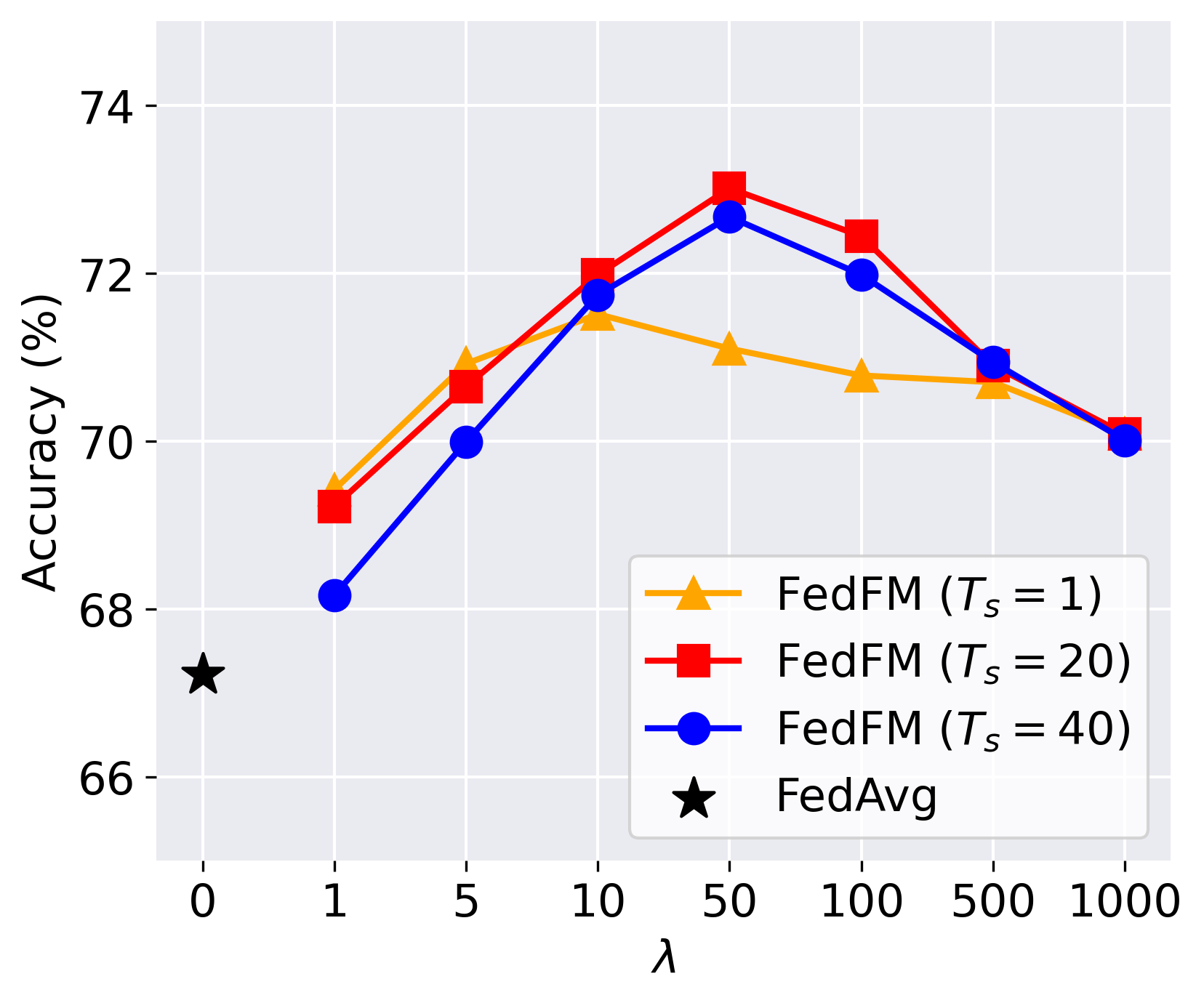}%
\label{fig:abl_lamt}}
\caption{Ablation study. (a) shows that FM consistently brings performance gain to four methods. (b) shows that CG significantly improve the effectiveness of FM. (c) shows that $T_s=20$ and $\lambda=10 \sim 100$ is roughly an optimal solution.}
\label{fig:ablation}
\end{figure*}

\subsubsection{Effects of global anchors aggregation manner}
\label{sssct:aggregation_manner}

Here, we show that FedFM can still achieve great performance without uploading clients' category distributions as discussed in Section~\ref{sssec:privacy}. We compare two manners of global anchors aggregation, sample-number-based weighted aggregation and uniform aggregation. For the weighted aggregation, each category-wise global anchor is updated by weighted aggregating local anchors according to each client's number of data samples of the corresponding category. This aggregation manner might not be allowed for its requirement for uploading clients' category distributions. For the uniform aggregation, each category-wise global anchor is updated by uniform aggregating local anchors of the corresponding category, which relieves the above issue. As an experimental detail, for those categories where a client has no data sample, we adopt the corresponding global anchors as the local anchors for aggregation.

We conduct experiments under NIID-1 on CIFAR-10~\cite{cifar10} and present the results in Table~\ref{table:anchor_aggregation}. Experiments show that FedFM with uniform aggregation performs comparably to FedFM with weighted aggregation (only $0.02\%$ performance drop). The reason behind this could be that for each category, all clients' features are pushed to the same shared global anchor. As a result, all clients' local anchors of the same category are close to each other, making it similar between applying weighted aggregation and uniform aggregation.

\subsubsection{Modularity of Feature Matching}
One advantage of our anchor-based feature matching (FM) method is its modularity, that is, it can be combined with most existing methods. Fig.~\ref{fig:abl_modularity} shows the performances before and after feature matching (FM) with contrastive-guiding combined with several existing methods. Here, we take FedAvg~\cite{fedavg}, FedProx~\cite{fedprox}, FedNova~\cite{fednova} and MOON~\cite{moon} as example and conduct experiments under NIID-1 setting on CIFAR10~\cite{cifar10}. Note that the previous explored FedFM corresponds to FedAvg~\cite{fedavg} incorporated with FM. 

From the figure, we see that applying our FM consistently brings performance gain to these four methods, achieving $4.63\%$ higher accuracy than corresponding baselines on average. Note that all these methods with FM outperforms the state-of-the-art performance $69.91\%$ (SCAFFOLD~\cite{scaffold}).

\subsubsection{Effects of contrastive-guiding loss}
Fig.~\ref{fig:abl_cg} presents the performance of FedAvg~\cite{fedavg}, FedFM with $\ell_2$ loss and FedFM with contrastive-guiding (CG) loss under different heterogeneous levels. Note that smaller $\beta$ corresponds to more severe heterogeneous level. We see that i) the performance of all methods degrades as the heterogeneous level increases ($\beta$ decreases), which verifies that data heterogeneity affects the performance of FL. ii) Both FedFM $\ell_2$ and CG loss outperform baseline FedAvg~\cite{fedavg}, indicating that anchor-based feature matching brings performance improvement to standard FL. iii) FedFM with CG significantly enhances the performance compared with FedFM with $\ell_2$, which indicates the effectiveness of our proposed CG loss.

\subsubsection{Effects of the weight of feature matching loss $\lambda$}
Fig.~\ref{fig:abl_lamt} presents the relationship between the final results and weight of feature matching loss $\lambda$. Specifically, for each curve in the figure, the launching round $T_s$ of FedFM is fixed while the $\lambda$ is tuned in $\{1, 5, 10, 50, 100, 500, 1000\}$. Note that when $\lambda = 0$, FedFM reduces to FedAvg~\cite{fedavg}, which is denoted by a star. We see that i) applying feature matching brings performance gain over FedAvg~\cite{fedavg} for a wide range of $\lambda$, indicating the effectiveness of feature matching; ii) a moderate $\lambda$ ranging from $10 \sim 100$ tends to perform better.

\subsubsection{Effects of the round $T_s$ to launch FedFM}
Fig.~\ref{fig:abl_lamt} presents the relationship between the final results and launching round $T_s$ of FedFM. Specifically, for each fixed $\lambda$, we compare the performance of three different $T_s \in \{1, 20, 40\}$. We see that a moderate $T_s=20$ performs the best or comparably in most cases. This is reasonable since at the initial rounds of FL, the established anchors are less representative and still in drastic change, which makes such feature matching less effective.

We also explore the effects of the number of epochs of local model training and the performance under partial client participation scenario in Table VII and VIII in appendix.

\section{Conclusion}

Facing statistical data heterogeneity, there are two unsatisfying phenomena in feature space for existing federated learning methods. Motivated by this, we propose a novel anchor-based federated feature matching (FedFM) method, which utilizes shared anchors to guide feature learning at multiple local models, promoting a consistent feature space. Tackling the theoretical challenge of varying objective function, we prove the convergence of FedFM. For more precise guiding, we further propose a novel contrastive-guiding (CG) loss, which guides the feature of each sample to match with the corresponding anchor while keeping far away from non-corresponding anchors. We propose a more efficient and flexible variant of FedFM, FedFM-Lite, which is capable of communicating anchors and models at different frequency. Experiments show that FedFM with CG (and FedFM-Lite) consistently outperform state-of-the-art methods.


\bibliographystyle{IEEEtran}
\bibliography{egbib}

\begin{thebibliography}{10}
\providecommand{\url}[1]{#1}
\csname url@samestyle\endcsname
\providecommand{\newblock}{\relax}
\providecommand{\bibinfo}[2]{#2}
\providecommand{\BIBentrySTDinterwordspacing}{\spaceskip=0pt\relax}
\providecommand{\BIBentryALTinterwordstretchfactor}{4}
\providecommand{\BIBentryALTinterwordspacing}{\spaceskip=\fontdimen2\font plus
\BIBentryALTinterwordstretchfactor\fontdimen3\font minus
  \fontdimen4\font\relax}
\providecommand{\BIBforeignlanguage}[2]{{%
\expandafter\ifx\csname l@#1\endcsname\relax
\typeout{** WARNING: IEEEtran.bst: No hyphenation pattern has been}%
\typeout{** loaded for the language `#1'. Using the pattern for}%
\typeout{** the default language instead.}%
\else
\language=\csname l@#1\endcsname
\fi
#2}}
\providecommand{\BIBdecl}{\relax}
\BIBdecl

\bibitem{fedavg}
B.~McMahan, E.~Moore, D.~Ramage, S.~Hampson, and B.~A. y~Arcas,
  ``Communication-efficient learning of deep networks from decentralized
  data,'' in \emph{Artificial intelligence and statistics}.\hskip 1em plus
  0.5em minus 0.4em\relax PMLR, 2017, pp. 1273--1282.

\bibitem{gafni2022federated}
T.~Gafni, N.~Shlezinger, K.~Cohen, Y.~C. Eldar, and H.~V. Poor, ``Federated
  learning: A signal processing perspective,'' \emph{IEEE Signal Processing
  Magazine}, vol.~39, no.~3, pp. 14--41, 2022.

\bibitem{advances}
P.~Kairouz, H.~B. McMahan, B.~Avent, A.~Bellet, M.~Bennis, A.~N. Bhagoji,
  K.~Bonawitz, Z.~Charles, G.~Cormode, R.~Cummings \emph{et~al.}, ``Advances
  and open problems in federated learning,'' \emph{Foundations and
  Trends{\textregistered} in Machine Learning}, vol.~14, no. 1--2, pp. 1--210,
  2021.

\bibitem{li_survey}
T.~Li, A.~K. Sahu, A.~Talwalkar, and V.~Smith, ``Federated learning:
  Challenges, methods, and future directions,'' \emph{IEEE Signal Processing
  Magazine}, vol.~37, no.~3, pp. 50--60, 2020.

\bibitem{smith2017federated}
V.~Smith, C.-K. Chiang, M.~Sanjabi, and A.~S. Talwalkar, ``Federated multi-task
  learning,'' \emph{Advances in neural information processing systems},
  vol.~30, 2017.

\bibitem{fedprox}
T.~Li, A.~K. Sahu, M.~Zaheer, M.~Sanjabi, A.~Talwalkar, and V.~Smith,
  ``Federated optimization in heterogeneous networks,'' \emph{Proceedings of
  Machine Learning and Systems}, vol.~2, pp. 429--450, 2020.

\bibitem{scaffold}
S.~P. Karimireddy, S.~Kale, M.~Mohri, S.~Reddi, S.~Stich, and A.~T. Suresh,
  ``Scaffold: Stochastic controlled averaging for federated learning,'' in
  \emph{International Conference on Machine Learning}.\hskip 1em plus 0.5em
  minus 0.4em\relax PMLR, 2020, pp. 5132--5143.

\bibitem{fedvisual}
T.-M.~H. Hsu, H.~Qi, and M.~Brown, ``Federated visual classification with
  real-world data distribution,'' in \emph{European Conference on Computer
  Vision}.\hskip 1em plus 0.5em minus 0.4em\relax Springer, 2020, pp. 76--92.

\bibitem{measuring}
{Hsu, Tzu-Ming Harry and Qi, Hang and Brown, Matthew}, ``Measuring the effects
  of non-identical data distribution for federated visual classification,''
  \emph{arXiv preprint arXiv:1909.06335}, 2019.

\bibitem{hard2018federated}
A.~Hard, K.~Rao, R.~Mathews, S.~Ramaswamy, F.~Beaufays, S.~Augenstein,
  H.~Eichner, C.~Kiddon, and D.~Ramage, ``Federated learning for mobile
  keyboard prediction,'' \emph{arXiv preprint arXiv:1811.03604}, 2018.

\bibitem{8683546}
D.~Leroy, A.~Coucke, T.~Lavril, T.~Gisselbrecht, and J.~Dureau, ``Federated
  learning for keyword spotting,'' in \emph{ICASSP 2019 - 2019 IEEE
  International Conference on Acoustics, Speech and Signal Processing
  (ICASSP)}, 2019, pp. 6341--6345.

\bibitem{abdulrahman2020survey}
S.~AbdulRahman, H.~Tout, H.~Ould-Slimane, A.~Mourad, C.~Talhi, and M.~Guizani,
  ``A survey on federated learning: The journey from centralized to distributed
  on-site learning and beyond,'' \emph{IEEE Internet of Things Journal},
  vol.~8, no.~7, pp. 5476--5497, 2020.

\bibitem{wangsurvey}
J.~Wang, Z.~Charles, Z.~Xu, G.~Joshi, H.~B. McMahan, M.~Al-Shedivat, G.~Andrew,
  S.~Avestimehr, K.~Daly, D.~Data \emph{et~al.}, ``A field guide to federated
  optimization,'' \emph{arXiv preprint arXiv:2107.06917}, 2021.

\bibitem{convergence_noniid}
X.~Li, K.~Huang, W.~Yang, S.~Wang, and Z.~Zhang, ``On the convergence of fedavg
  on non-iid data,'' in \emph{International Conference on Learning
  Representations}, 2019.

\bibitem{fednoniid}
Y.~Zhao, M.~Li, L.~Lai, N.~Suda, D.~Civin, and V.~Chandra, ``Federated learning
  with non-iid data,'' \emph{arXiv preprint arXiv:1806.00582}, 2018.

\bibitem{feddyn}
D.~A.~E. Acar, Y.~Zhao, R.~Matas, M.~Mattina, P.~Whatmough, and V.~Saligrama,
  ``Federated learning based on dynamic regularization,'' in
  \emph{International Conference on Learning Representations}, 2020.

\bibitem{tsne}
L.~Van~der Maaten and G.~Hinton, ``Visualizing data using t-sne.''
  \emph{Journal of machine learning research}, vol.~9, no.~11, 2008.

\bibitem{fednova}
J.~Wang, Q.~Liu, H.~Liang, G.~Joshi, and H.~V. Poor, ``A novel framework for
  the analysis and design of heterogeneous federated learning,'' \emph{IEEE
  Transactions on Signal Processing}, vol.~69, pp. 5234--5249, 2021.

\bibitem{fedavgm}
T.-M.~H. Hsu, H.~Qi, and M.~Brown, ``Measuring the effects of non-identical
  data distribution for federated visual classification,'' \emph{arXiv preprint
  arXiv:1909.06335}, 2019.

\bibitem{moon}
Q.~Li, B.~He, and D.~Song, ``Model-contrastive federated learning,'' in
  \emph{Proceedings of the IEEE/CVF Conference on Computer Vision and Pattern
  Recognition}, 2021, pp. 10\,713--10\,722.

\bibitem{cifar10}
A.~Krizhevsky, G.~Hinton \emph{et~al.}, ``Learning multiple layers of features
  from tiny images,'' 2009.

\bibitem{darlow2018cinic}
L.~N. Darlow, E.~J. Crowley, A.~Antoniou, and A.~J. Storkey, ``Cinic-10 is not
  imagenet or cifar-10,'' \emph{arXiv preprint arXiv:1810.03505}, 2018.

\bibitem{fedmedical}
G.~A. Kaissis, M.~R. Makowski, D.~R{\"u}ckert, and R.~F. Braren, ``Secure,
  privacy-preserving and federated machine learning in medical imaging,''
  \emph{Nature Machine Intelligence}, vol.~2, no.~6, pp. 305--311, 2020.

\bibitem{fedhealth}
Y.~Chen, X.~Qin, J.~Wang, C.~Yu, and W.~Gao, ``Fedhealth: A federated transfer
  learning framework for wearable healthcare,'' \emph{IEEE Intelligent
  Systems}, vol.~35, no.~4, pp. 83--93, 2020.

\bibitem{9832948}
Z.~Chen, C.~Yang, M.~Zhu, Z.~Peng, and Y.~Yuan, ``Personalized
  retrogress-resilient federated learning towards imbalanced medical data,''
  \emph{IEEE Transactions on Medical Imaging}, pp. 1--1, 2022.

\bibitem{vrlsgd}
X.~Liang, S.~Shen, J.~Liu, Z.~Pan, E.~Chen, and Y.~Cheng, ``Variance reduced
  local sgd with lower communication complexity,'' \emph{arXiv preprint
  arXiv:1912.12844}, 2019.

\bibitem{fedgen}
Z.~Zhu, J.~Hong, and J.~Zhou, ``Data-free knowledge distillation for
  heterogeneous federated learning,'' in \emph{International Conference on
  Machine Learning}.\hskip 1em plus 0.5em minus 0.4em\relax PMLR, 2021, pp.
  12\,878--12\,889.

\bibitem{feddf}
T.~Lin, L.~Kong, S.~U. Stich, and M.~Jaggi, ``Ensemble distillation for robust
  model fusion in federated learning,'' \emph{Advances in Neural Information
  Processing Systems}, vol.~33, pp. 2351--2363, 2020.

\bibitem{fedftg}
L.~Zhang, L.~Shen, L.~Ding, D.~Tao, and L.-Y. Duan, ``Fine-tuning global model
  via data-free knowledge distillation for non-iid federated learning,'' in
  \emph{Proceedings of the IEEE/CVF Conference on Computer Vision and Pattern
  Recognition}, 2022, pp. 10\,174--10\,183.

\bibitem{fedrep}
L.~Collins, H.~Hassani, A.~Mokhtari, and S.~Shakkottai, ``Exploiting shared
  representations for personalized federated learning,'' in \emph{International
  Conference on Machine Learning}.\hskip 1em plus 0.5em minus 0.4em\relax PMLR,
  2021, pp. 2089--2099.

\bibitem{fedamp}
Y.~Huang, L.~Chu, Z.~Zhou, L.~Wang, J.~Liu, J.~Pei, and Y.~Zhang,
  ``Personalized cross-silo federated learning on non-iid data,'' in
  \emph{Proceedings of the AAAI Conference on Artificial Intelligence},
  vol.~35, no.~9, 2021, pp. 7865--7873.

\bibitem{pfedme}
C.~T~Dinh, N.~Tran, and J.~Nguyen, ``Personalized federated learning with
  moreau envelopes,'' \emph{Advances in Neural Information Processing Systems},
  vol.~33, pp. 21\,394--21\,405, 2020.

\bibitem{per_ml_2}
A.~Fallah, A.~Mokhtari, and A.~Ozdaglar, ``Personalized federated learning with
  theoretical guarantees: A model-agnostic meta-learning approach,''
  \emph{Advances in Neural Information Processing Systems}, vol.~33, pp.
  3557--3568, 2020.

\bibitem{tan2021fedproto}
Y.~Tan, G.~Long, L.~Liu, T.~Zhou, Q.~Lu, J.~Jiang, and C.~Zhang, ``Fedproto:
  Federated prototype learning over heterogeneous devices,'' \emph{arXiv
  preprint arXiv:2105.00243}, 2021.

\bibitem{resnet}
K.~He, X.~Zhang, S.~Ren, and J.~Sun, ``Deep residual learning for image
  recognition,'' in \emph{Proceedings of the IEEE conference on computer vision
  and pattern recognition}, 2016, pp. 770--778.

\bibitem{feature_inversion}
N.~Zhao, Z.~Wu, R.~W. Lau, and S.~Lin, ``What makes instance discrimination
  good for transfer learning?'' in \emph{International Conference on Learning
  Representations}, 2020.

\bibitem{nofear}
M.~Luo, F.~Chen, D.~Hu, Y.~Zhang, J.~Liang, and J.~Feng, ``No fear of
  heterogeneity: Classifier calibration for federated learning with non-iid
  data,'' \emph{Advances in Neural Information Processing Systems}, vol.~34,
  2021.

\bibitem{bonawitz2017practical}
K.~Bonawitz, V.~Ivanov, B.~Kreuter, A.~Marcedone, H.~B. McMahan, S.~Patel,
  D.~Ramage, A.~Segal, and K.~Seth, ``Practical secure aggregation for
  privacy-preserving machine learning,'' in \emph{proceedings of the 2017 ACM
  SIGSAC Conference on Computer and Communications Security}, 2017, pp.
  1175--1191.

\bibitem{reddi2020adaptive}
S.~J. Reddi, Z.~Charles, M.~Zaheer, Z.~Garrett, K.~Rush, J.~Kone{\v{c}}n{\`y},
  S.~Kumar, and H.~B. McMahan, ``Adaptive federated optimization,'' in
  \emph{International Conference on Learning Representations}, 2020.

\bibitem{fedma}
\BIBentryALTinterwordspacing
H.~Wang, M.~Yurochkin, Y.~Sun, D.~Papailiopoulos, and Y.~Khazaeni, ``Federated
  learning with matched averaging,'' in \emph{International Conference on
  Learning Representations}, 2020. [Online]. Available:
  \url{https://openreview.net/forum?id=BkluqlSFDS}
\BIBentrySTDinterwordspacing

\bibitem{bayesian}
M.~Yurochkin, M.~Agarwal, S.~Ghosh, K.~Greenewald, N.~Hoang, and Y.~Khazaeni,
  ``Bayesian nonparametric federated learning of neural networks,'' in
  \emph{International Conference on Machine Learning}.\hskip 1em plus 0.5em
  minus 0.4em\relax PMLR, 2019, pp. 7252--7261.

\bibitem{rousseeuw1987silhouettes}
P.~J. Rousseeuw, ``Silhouettes: a graphical aid to the interpretation and
  validation of cluster analysis,'' \emph{Journal of computational and applied
  mathematics}, vol.~20, pp. 53--65, 1987.

\bibitem{lloyd1982least}
S.~Lloyd, ``Least squares quantization in pcm,'' \emph{IEEE transactions on
  information theory}, vol.~28, no.~2, pp. 129--137, 1982.

\bibitem{pedregosa2011scikit}
F.~Pedregosa, G.~Varoquaux, A.~Gramfort, V.~Michel, B.~Thirion, O.~Grisel,
  M.~Blondel, P.~Prettenhofer, R.~Weiss, V.~Dubourg \emph{et~al.},
  ``Scikit-learn: Machine learning in python,'' \emph{the Journal of machine
  Learning research}, vol.~12, pp. 2825--2830, 2011.

\end{thebibliography}
\end{document}